\documentclass{article} 
\usepackage{arxiv}

\usepackage[american]{babel}

\usepackage{natbib} 
\usepackage{mathtools} 
\usepackage{booktabs} 
\usepackage{tikz} 
\usepackage{graphicx}

\usepackage{amsmath,amsfonts,amsthm}
\newtheorem{assumption}{Assumption}
\newtheorem{definition}{Definition}
\newtheorem{theorem}{Theorem}
\newtheorem{corollary}{Corollary}
\newtheorem{lemma}{Lemma}

\DeclareMathOperator*{\argmax}{arg\,max}

\usepackage{subcaption}
\newcommand\myrowlabel[1]{%
  \rotatebox[origin=c]{90}{#1}%
}

\usepackage{algpseudocode}
\usepackage{algorithm}
\usepackage{setspace}

\newcommand*\diff{\mathop{}\!\mathrm{d}}

\date{}

\author{ Mathieu Godbout \\
    Institut Intelligence et Données (IID)\\
	Department of Computer Science\\
	Université Laval\\
	\texttt{mathieu.godbout.3@ulaval.ca}
	\And
	Maxime Heuillet \\
	Institut Intelligence et Données (IID)\\
	Department of Computer Science\\
	Université Laval\\
	\texttt{maxime.heuillet.1@ulaval.ca}
	\And
	Sharath Chandra Raparthy \\
	Quebec Artificial Intelligence Institute (Mila)\\
	Department of Computer Science and Operations Research\\
	Université de Montréal\\
	\texttt{raparths@mila.quebec}
    \And
    Rupali Bhati \\
    Institut Intelligence et Données (IID)\\
	Department of Computer Science\\
	Université Laval\\
	\texttt{rupali.bhati.1@ulaval.ca}
	\And
	Audrey Durand\\
	CIFAR AI Chair\\
	Institut Intelligence et Données (IID)\\
	Department of Computer Science\\
	Université Laval\\
	\texttt{audrey.durand@ift.ulaval.ca}
}

\title{ACReL: Adversarial Conditional value-at-risk Reinforcement Learning}

\begin{document}
\maketitle

\begin{abstract}
In the classical Reinforcement Learning (RL) setting, one aims to find a policy that maximizes its expected return.
This objective may be inappropriate in safety-critical domains such as healthcare or autonomous driving, where intrinsic uncertainties due to stochastic policies and environment variability may lead to catastrophic failures.
This can be addressed by using the Conditional-Value-at-Risk (CVaR) objective to instill risk-aversion in learned policies.
In this paper, we propose Adversarial Cvar Reinforcement Learning (ACReL), a novel adversarial meta-algorithm to optimize the CVaR objective in RL.
ACReL is based on a max-min between a policy player and a learned adversary that perturbs the policy player's state transitions given a finite budget. 
We prove that, the closer the players are to the game's equilibrium point, the closer the learned policy is to the CVaR-optimal one with a risk tolerance explicitly related to the adversary's budget.
We provide a gradient-based training procedure to solve the proposed game by formulating it as a Stackelberg game, enabling the use of deep RL architectures and training algorithms. 
Empirical experiments show that ACReL matches a CVaR RL state-of-the-art baseline for retrieving CVaR optimal policies, while also benefiting from theoretical guarantees.
\end{abstract}

\section{Introduction}
\label{sec:introduction}

Reinforcement Learning (RL)~\citep{sutton1998introduction} is a branch of machine learning where the learner (agent) must learn how to behave in an environment by trial and error. 
In the classical RL setting, the objective is to find agents that maximize their expected return.
This expectation maximization objective has led to impressive successes in domains like videogames~\citep{vinyals2019grandmaster}, board games~\citep{silver2018general} or content recommendation~\citep{linucb}.
However, in safety-critical domains like healthcare, autonomous driving or financial planning, some erroneous actions may lead to disastrous consequences.
In healthcare for instance, an RL agent may be in charge of designing the shortest path to an organ for surgery.
In this task, some paths may be shorter but at the same time may be risk endangering the patient as they are too close to an artery, a nerve or a critical region of the brain~\citep{pathRLsurgery}. 
Automation in this context and other safety-critical domains can only come if the agent is able to successfully reach its goal while avoiding the most risky actions~\citep{gottesman2019guidelines}. 
Therefore, the classical RL approach of naively maximizing the expected return is not appropriate in such scenarios.

This has motivated the community to design risk-sensitive agents trained to take into account the uncertainty over their return, with special attention on catastrophic events. 
This can be achieved by including a risk measure~\citep{artzner1999coherent} in the learning objective. 
A common risk measure is the Conditional Value-at-Risk (CVaR$_\alpha$), defined as the expectation over the worst $\alpha$-quantile of a distribution. 
Lowering the value of $\alpha$ increases the risk-aversion.
The search for CVaR-optimal policies, usually referred to as CVaR RL, is typically achieved with distributional RL~\citep{dabney2018implicit,keramati2020being,singh2020improving,schubert2021automatic}. 
However, these strategies lack theoretical guarantees on convergence to risk-sensitive objectives~\citep{dabney2018implicit}.

Inspired by how the game-theoretic perspective has helped advance other fields like Generative Adversarial Networks (GANs)~\citep{berthelot2017began,goodfellow2020generative}, we address the lack of CVaR RL strategies with theoretical optimality guarantees by borrowing tools from game theory.
More precisely, we suggest that learning the CVaR optimal policy in a risk-averse RL setting can be cast as a max-min game~\citep{neumann1928theorie} between the RL agent and a learned adversary. 
In the resulting setting that we call Adversarial Cvar Reinforcement Learning (ACReL), a learning agent aiming at maximizing its reward collection is perturbed by an adversary, whose goal is to minimize the rewards obtained by the agent. 
Our contributions are the following:
\begin{itemize}
    \item we propose ACReL, a max-min game formulation of CVaR RL with learnable agent and adversarial policies;
    \item we provide a theoretical analysis showing that the game equilibrium yields CVaR-optimal policies;
    \item we provide a theoretical analysis of a gradient-based algorithm to solve the proposed game.
\end{itemize}

The rest of the paper is organized as follows.
We first introduce the necessary background and notation (Section~\ref{sec:background}).
We then formulate the ACReL setting alongside its relevant properties allowing to conduct a theoretical convergence analysis linking the game equilibrium point to a CVaR-optimal RL policy (Section~\ref{sec:game}).
This leads us to consider a gradient-based algorithm leveraging the Stackelberg game formulation, for which we provide a theoretical convergence analysis (Section~\ref{sec:stackelberg}). 
We present empirical results showing that the proposed approach matches the performance of a state-of-the-art risk-sensitive baseline in finding CVaR-optimal RL policies (Section~\ref{sec:experiments}).
Lastly, we provide an overview of the related work (Section~\ref{sec:related_work}).

\section{Background and Notation}
\label{sec:background}

The most common RL problem formulation is under the Markov Decision Process (MDP) framework.
An MDP is represented by a tuple $\langle \mathcal{S}, \mathcal{A}, \mathcal{P}, \mathcal{R}, \gamma  \rangle$, where $\mathcal{S}$ is the state space, $\mathcal{A}$ is the action space, $\mathcal{P}(\cdot | s, a)$ is the transition kernel which governs the evolution of the system, $\mathcal{R}$ is the reward function for a given state-action pair, and $\gamma \in [0, 1)$ is a discount factor. 
On each time step $t$, the learning agent evolving in the described environment observes the current state $s_t\in\mathcal S$, takes an action $a_t\in\mathcal A$, transitions in the next state $s_{t+1} \sim \mathcal P(s_t, a_t)$, and observes the associated reward $r_{t+1}\sim\mathcal R(s_t, a_t, s_{t+1})$.
The behavior of the agent is represented by a policy $\pi(\cdot | s)$ mapping any state $s\in\mathcal S$ to a probability distribution over actions, such that $a_t \sim \pi(s_t)$.

Selected actions not only generate immediate rewards but also influence future rewards because they rule the next state the agent will end up in.
Therefore, for any given state $s \in \mathcal{S}$, the performance of a policy $\pi$ is measured by balancing immediate and distant rewards using the discounted return:
\begin{equation}
    \mathcal{J}(\pi) :=  \sum_{t = 0}^{\infty} \gamma^t r_t.
    \label{eq:J_random_return}
\end{equation}
Due to the possible stochasticity in the reward function, the state transitions, or noise in the policy, the discounted return $\mathcal{J}(\pi)$ is accurately defined as a random variable.
The classical objective in RL is to find the policy $\pi^*$ that maximizes its expected discounted return $\pi^*= \argmax_\pi \mathbb{E}[\mathcal{J}(\pi)]$.

\paragraph{Conditional-Value-at-Risk}

Let $Z$ be a random variable with a bounded expectation ($\mathbb{E}[Z] < \infty$) and cumulative distribution function $F(z):= \mathbb{P}(Z\leq z)$.
The Value-at-Risk ($\text{VaR}_\alpha$) at confidence level $\alpha \in (0, 1]$ represents the worst $\alpha$-quantile of $Z$, i.e., $\text{VaR}_\alpha(Z):= \min_z\{z | F(z) \geq \alpha\}$.
Analogously, the Conditional-Value-at-Risk~\citep{artzner1999coherent} at confidence level $\alpha$ for a continuous distribution $Z$ is defined as
\begin{equation}
\label{eq:CVaR_continuous}
\text{CVaR}_\alpha [Z] := \mathbb{E} \left[z \,|\, z \leq \text{VaR}_\alpha(Z) \right].
\end{equation}
Intuitively, the CVaR$_\alpha[Z]$ can be viewed as the mean over the $\alpha$-quantile worst values of $Z$.
Since $Z$ represents discounted returns in our case, the worst values of $Z$ can be seen as the ones where most risk has been incurred.
This last interpretation is particularly attractive, as it makes the CVaR$_\alpha$ easy to understand for non-experts who might be involved in the design of any risk-sensitive model in safety-critical domains. 

\paragraph{CVaR Reinforcement Learning}
To measure the level of risk associated with a policy $\pi$, the CVaR$_\alpha$ measure (Eq.~\ref{eq:CVaR_continuous}) can be applied to the discounted return $\mathcal{J}(\pi)$ (Eq.~\ref{eq:J_random_return}).
Optimizing for this yields the CVaR RL objective 
\begin{equation}
\label{eq:CVaR_RL}
\pi^* = \argmax_{\pi} \text{CVaR}_\alpha \left[\mathcal{J}(\pi)\right].
\end{equation}
This objective, which differs from the classical expectation maximization goal, can be viewed as finding a risk-sensitive policy.
Indeed, by aiming for the optimal CVaR$_\alpha$ return, the optimal policy $\pi^*$ is maximizing the expectation over the $\alpha$-tail worst returns.
Since the expectation is now taken over a small, disadvantageous subset of returns rather than over all of them, $\pi^*$ is more sensible towards avoiding the probability of large negative returns.
In practice, this means that optimizing for the CVaR RL objective (Eq.~\ref{eq:CVaR_RL}) will produce policies that accept to reduce their expected performance, so long as it means they avoid catastrophic trajectories in return. 
In other words, they will reduce uncertainty at the expense of performance.

\section{ACReL}
\label{sec:game}

Before diving into the details of the proposed adversarial game, let us first present the necessary assumption on the target MDP for the game to be applicable.

\begin{assumption}
\label{ass:stochastic}
State transitions $\mathcal{P}$ are stochastic, they can be accessed, and their non-zero values can be arbitrarily modified.
\end{assumption}

Stochasticity is required because it is precisely upon state transitions uncertainties that the adversary will be acting.
State transitions must also be accessible to allow adversarial actions to be registered.
The accessibility requirement, albeit more restrictive, is likely to be met in artificial environments over which the user has almost complete control, in the Sim2Real~\citep{peng2018sim} context for instance.
We however emphasize that accessibility is only needed to characterize the game's equilibrium as theoretically CVaR optimal.
This means that our game framework may still be used in practice with different adversarial disturbances, only at the cost of losing the presented theoretical guarantees on the game's solution. 

\subsection{Game Setting}

The goal is to learn a risk-sensitive policy with respect to the CVaR$_\alpha$ risk measure.
To achieve this, we leverage the perturbation model from~\citet{chow2015risk} to design a max-min game where an agent tries to learn a policy $\pi_\theta$ to maximize its return while an adversary tries to learn a perturbation policy $\nu_\omega$ that instead minimizes the agent's return, where $\theta$ and $\omega$ respectively denote the parameters of the agent and adversarial policies.

We note that, due to the time-inconsistent nature of CVaR in sequential problems~\citep{gagne2021two}, it is well known that the optimal policy $\pi^*$ for the CVaR RL objective can be history-dependent~\citep{shapiro2014lectures}.
We hereby follow previous work on CVaR RL~\citep{keramati2020being,dabney2018implicit} and limit our analysis to stationary, history independent policies which can be suboptimal but typically achieve high CVaR nonetheless in practice.

\paragraph{The adversary}
Let $\mathcal{P}_t = \mathcal{P}(\cdot \, |\, s_t, a_t)$ denote time step transition dynamics given a trajectory $\tau = \{ (s_t,a_t,r_t)\}_{t=1}^T$ performed by the agent, where $T$ is the duration of the trajectory.
At time step $t$, the adversary can access and perturb $\mathcal{P}_t$ with a \textit{positive multiplicative probability perturbation} $\delta_t$.
This yields perturbed transitions $\hat{\mathcal{P}}_t=\mathcal{P}_t \circ \delta_t$, where $\circ$ denotes the Hadamard (element-wise) product.
To be admissible, a perturbation $\delta_t$ must generate a valid transition kernel, i.e. $\hat{\mathcal{P}}_t(s) \geq 0$ for all state $s\in\mathcal{S}$, and $\int_\mathcal{S} \hat{\mathcal{P}}_t(s)\diff s = 1$.
This constrains the perturbations to states with non-zero transition probability in $\mathcal P_t$, i.e. the adversary is only allowed to modify the distribution over next states that were already reachable.
At the beginning of a trajectory, the adversary is given an initial perturbation budget $\eta_1$, which gets updated after each state transition.
On time step $t$, the adversary is constrained by the remaining budget $\eta_t$, i.e. they must enforce that $\delta_t(s) \leq \eta_t$ for all $s \in \mathcal{S}$.
To ease notation, we hereafter refer to $\eta=\eta_1$ as the adversary budget.
The budget constraint is implemented over entire trajectories, i.e. $\delta_1(s_2)\cdot...\cdot\delta_{T-1}(s_T) \leq \eta$.
The adversary therefore needs to maximize the impairment to the agent without wasting their budget.

\begin{figure}
    \centering
    \includegraphics[width=0.4\textwidth]{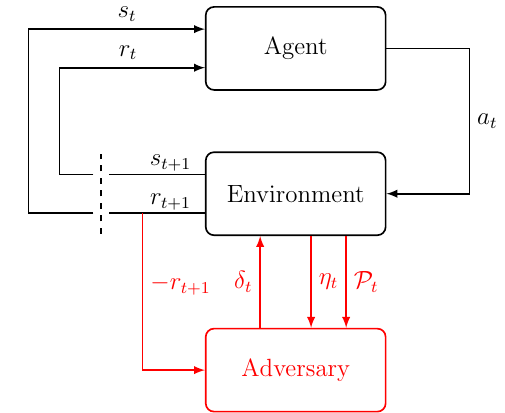}
    \caption{ACReL interactions dynamic.
            Components that deviate from standard RL are highlighted in red.}
    \label{fig:CARL_loop}
\end{figure}

\paragraph{Game dynamics}
The game between the agent and the adversary goes as follows. On each time step $t$, the agent observes their state $s_t$ and selects an action $a_t$.
The adversary then observes their state $s_t' = (\mathcal{P}_t, \eta_t)$ given by the non-perturbed transition kernel $\mathcal P_t$ of the agent and the current budget $\eta_t$. Using this information, they select an adversarial action (perturbation) such that $\delta_t(s) \leq \eta_t~\forall s\in\mathcal S$, generating the perturbed transitions $\hat{\mathcal P}_t$. The agent then transitions into the next state $s_{t+1}\sim \hat{\mathcal P}_t$, a reward $r_{t+1}\sim \mathcal R(s_t, a_t, s_{t+1})$ is generated, and the remaining budget is updated according to the next state's corresponding perturbation via $\eta_{t+1}=\eta_t/\delta_t(s_{t+1})$. 
The agent obtains the reward $r_{t+1}$, while the adversary obtains the reward $-r_{t+1}$.
Figure~\ref{fig:CARL_loop} shows an overview of the ACReL game dynamics.

Note that, since the perturbation budget is only affected by the next encountered state $s_{t+1} \sim \hat{\mathcal{P}_t}$, the budget decreases only if the adversary increased the probability of sampling $s_{t+1}$, i.e. if $\delta_t(s_{t+1}) > 1$.
This also implies that if $\delta_t(s_{t+1}) < 1$, the budget will increase.
We emphasize that the adversary may not however deliberately aim to increase their budget.
Indeed, since any budget increase is the result of the next state being sampled despite the adversary decreasing its probability, there also has to be an equivalent likelihood that the next sampled state had its probability increased and would result in a budget decrease instead.

\paragraph{Max-min game}
The general idea of this work is that an adversarial policy $\nu$ can be learned to select perturbations that respect the budget constraint, attempting to minimize the rewards collected by following policy $\pi$.
Combining the agent and adversary naturally leads to the max-min game
\begin{equation}
\label{eq:tpzs_CARL}
\max_\pi \min_{\nu} \mathbb{E} \left[\mathcal{J}^\eta(\pi, \nu)\right],
\end{equation}
where $\mathbb{E}[\mathcal{J}^\eta(\pi, \nu)]$ denotes the expected return collected under policy $\pi$ perturbed by an adversarial policy $\nu$ with initial budget $\eta$.
Interestingly enough, this formulation implies that the goal of each player is to respectively maximize and minimize the expected return, retrieving the classical risk-neutral RL objective.
This means that any standard RL algorithm may be applied almost out-of-the-box to this game.
Moreover, the embedded flexibility of the game also allows for the underlying MDP to have continuous action and/or state spaces.


\subsection{Game Solution}

The solution to the game \eqref{eq:tpzs_CARL} is given by its \textit{Minimax Equilibrium}.

\begin{definition}
(Minimax Equilibrium).
A Minimax Equilibrium $(\pi^*, \nu^*)$ is defined as a point where
\begin{equation}
\label{eq:minimax_eq}
\mathbb{E}\left[\mathcal{J}^\eta(\pi, \nu^*) \right] \leq \mathbb{E}\left[\mathcal{J}^\eta(\pi^*, \nu^*) \right] \leq \mathbb{E}\left[\mathcal{J}^\eta(\pi^*, \nu) \right],
\end{equation}
for which the inequalities hold for any other agent $\pi$ or adversary $\nu$.
\end{definition}

At this equilibrium point, we have the following result.
\begin{lemma}
\label{thm:equilibrium_cvar}
Suppose $\pi_\theta$ and $\nu_\omega$ are expressive enough player parametrizations, meaning that they can represent the optimal players $\pi^*$ and $\nu^*$.
Then, at the Minimax Equilibrium $(\pi^*, \nu^*)$, the agent policy $\pi^*$ is {\normalfont CVaR}$_\alpha$ optimal for $\alpha=\frac{1}{\eta}$, where $\eta$ is the adversarial perturbation budget:
\begin{equation*}
    \max_\pi \min_{\nu} \mathbb{E} \left[\mathcal{J}^\eta(\pi, \nu)\right] =  \max_\pi \text{\normalfont CVaR}_{\frac{1}{\eta}} \left[ \mathcal{J}(\pi)\right].
\end{equation*}
\end{lemma}
\begin{proof}
Proposition~1 from~\citet{chow2015risk} established that
\begin{equation*}
    \inf_{\nu'} \mathbb{E}\left[\mathcal{J}^\eta(\pi, \nu') \right] = \text{CVaR}_{\frac{1}{\eta}} \left[\mathcal{J}(\pi)\right].
\end{equation*}
The result follows by observing that the $\inf$ operation will return an adversary in the search space and then taking the $\max$ over the policy search space.
\end{proof}
The result in Lemma~\ref{thm:equilibrium_cvar} is satisfying in two ways.
First, it directly translates the ACReL framework into risk-sensitivity for the agent policy, even though the agent and adversary both have classical risk-neutral objectives.
Second, it establishes that the $\alpha$ confidence interval in the CVaR$_\alpha$ objective is directly linked to the perturbation budget $\eta$.
Accordingly, in a real-world scenario where one wishes to optimize for a given $\alpha$-quantile, setting the initial adversarial budget to $\eta=\frac{1}{\alpha}$ will yield the desired risk-sensitive policy at the equilibrium point.

Given the parametrized ACReL setting where both the agent and adversarial policies are learned rather than exactly computed, we are particularly concerned with the notion of approximate equilibrium, which is more likely to be encountered in practice.
\begin{definition}
($\varepsilon$-Minimax Equilibrium).
Consider $V^* := \mathbb{E}\left[\mathcal{J}^\eta(\pi^*, \nu^*) \right]$, the value at the Minimax Equilibrium $(\pi^*, \nu^*)$.
An $\varepsilon$-Minimax equilibrium is a pair of players $(\pi', \nu')$ where
\begin{equation}
\label{eq:approx_minimax_eq}
\begin{split}
& V^* - \min_\nu \mathbb{E}\left[\mathcal{J}^\eta(\pi', \nu) \right] \leq \varepsilon\\
 \text{and} \quad  & \max_\pi \mathbb{E}\left[\mathcal{J}^\eta(\pi, \nu') \right] - V^* \leq \varepsilon.
\end{split}
\end{equation}
\end{definition}

Interestingly, this approximate equilibrium yields an approximately CVaR-optimal policy.
\begin{theorem}
\label{thm:approx_equilibrium_cvar}
Suppose $\pi_\theta$ and $\nu_\omega$ are expressive enough player parametrizations.
Then, at an $\varepsilon$-Minimax Equilibrium point $(\pi', \nu')$, the agent's policy $\pi'$ is at most $\varepsilon$-suboptimal:
\begin{equation*}
    \max_\pi\text{\normalfont CVaR}_{\frac{1}{\eta}}\left[\mathcal{J}(\pi)\right] - \text{\normalfont CVaR}_{\frac{1}{\eta}}\left[\mathcal{J}(\pi')\right] \leq \varepsilon.
\end{equation*}
\end{theorem}
\begin{proof}
Replacing the two terms on the LHS of Eq.~\ref{eq:approx_minimax_eq} by Lemma~\ref{thm:equilibrium_cvar} and Proposition~1 from~\citet{chow2015risk} respectively leads to the result.
\end{proof}

\section{Gradient-based Algorithm}
\label{sec:stackelberg}


Following the success of Deep RL to tackle challenging RL problems~\citep{arulkumaran2017deep}, we aim to leverage advanced (deep) representations for solving ACReL.
However, computing the solution of the max-min game (Eq.~\ref{eq:tpzs_CARL}) for players represented by neural networks is not trivial.
Indeed, the dependence between the rewards observed by each player makes the optimization objective non-stationary~\citep{fiez2019convergence}, hampering the convergence properties of vanilla gradient-based algorithms.
It is therefore necessary to take into account the game structure in the parameter updates of the agent and adversarial policies.
Although no gradient-based optimization updates specifically designed for max-min games like ACReL currently exists, such updates do exist for Stackelberg games~\citep{von2010Market}.
We therefore cast ACReL under the Stackelberg formulation to derive the desired gradient updates.

\subsection{Stackelberg Algorithm}

A two-player Stackelberg game~\citep{von2010Market} is an asymmetric game played between a leader $L$ and a follower $F$.
Both players aim to maximize their respective payoff functions $u_L(\theta_L, \theta_F)$ and $u_F(\theta_L, \theta_F)$, given their respective parameters $\theta_L$ and $\theta_F$. 
The particularity in this game is that $\theta_F$ always represents a best-response with respect to $\theta_L$.
Accordingly, the leader can then update its parameters $\theta_L$ by taking for granted that $\theta_F$ will be updated to remain optimal with respect to them.
This yields the following optimization problem for the leader
\begin{equation*}
\theta_L \in \argmax_\theta \left\{ u_L(\theta, \theta_F^{\star}) \, \, \Bigg | \, \, \theta_F^{\star} \in \argmax_{\theta_F} u_F(\theta, \theta_F) \right\}
\end{equation*}
and for the follower
\begin{equation*}
\theta_F \in \argmax_{\theta} u_F(\theta_L, \theta).
\end{equation*}
Note here that since $\theta_F^{\star}$ are always presumed to be optimal with respect to the leader, they may be seen as an implicit function taking as input the $\theta_L$, i.e. $\theta_F^{\star}=\theta_F^{\star}(\theta_L)$.
Thus, the optimization problem for the leader only depends on its parameters $\theta_L$ and the non-stationarity issue arising from the game objective may be circumvented by iteratively optimizing each player parameters independently.

The key to solving a Stackelberg game is to ensure that the follower learns much faster than the leader, so that its parameters always constitute a best response to the leader.
We can therefore simply alternate between solving the optimizing problem of each player, resulting in a simple yet effective algorithm.
Since max-min games such as ACReL are a special case of Stackelberg games where $u_L=-u_F$, we need only to define a leader and follower to apply the Stackelberg algorithm.
We select the agent as the leader and the adversary as the follower, but show in our convergence analysis (Section~\ref{subsec:conv_analysis}) that this choice can be reversed.

\subsection{Practical Implementation}

In practice, we adopt common relaxations of the Stackelberg algorithm (Section~\ref{sec:stackelberg}), which have proven effective with GANs~\citep{heusel2017gans} and model-based RL~\citep{rajeswaran2020game}.
First, we implement \textit{near optimal} updates of the follower by performing $K_\text{adv} > 1$ gradient steps of the adversary (follower) before updating the agent (leader).
Second, we use a first-order approximation of the joint gradient to update the leader rather than the computationally expensive exact update.
Algorithm~\ref{alg:algorithme} shows the resulting ACReL game.
Further details on the practical implementation can be found in Appendix~\ref{app:prac_algo_details}.

\begin{algorithm}[t!]
\setstretch{1.2}
    \caption{ACReL}
    \label{alg:algorithme}
    \begin{algorithmic}[1]
        \Require $\pi_\theta$ (agent policy), $\nu_\omega$ (adversarial policy), $\eta$ (budget), $K_\text{adv}$ (number of steps between adversary updates)
        \State $N=0$
        \While{not done}
        \State Initialize budget $\eta_1 = \eta$
        \State Initialize time step $t = 1$
        \State Agent observes initial state $s_1$
        \While{state $s_t$ is not terminal}
        \State Agent plays $a_t \sim \pi_\theta(s_t)$
        \State Adversary plays $\delta_t = \nu_\omega(\mathcal P_t, \eta_t)$
        \State Agent transitions to $s_{t+1} \sim \hat{\mathcal P}_t$
        \State Reward $r_{t+1} \sim \mathcal R(s_t, a_t, s_{t+1})$ is generated
        \State Budget is updated: $\eta_{t+1} = \eta_t/\delta_t(s_{t+1})$
        \State Move on to the next time step: $t = t + 1$
        \EndWhile
        \State Update $\theta$ if $N \mod (K_\text{adv} + 1) = 0$, otherwise $\omega$. 
        \State $N$ = $N$ + 1
        \EndWhile
    \end{algorithmic}
\end{algorithm}


\subsection{Convergence Analysis}
\label{subsec:conv_analysis}

Given the following assumption on the MDP, the Stackelberg algorithm (Section~\ref{sec:stackelberg}) converges to the equilibrium of the ACReL game (Eq.~\ref{eq:minimax_eq}).

\begin{assumption}
\label{ass:smooth_convex}
The reward function $\mathcal{R}$ and all transition functions $\mathcal{P}$ are both smooth and convex.
\end{assumption}
The smoothness assumption is likely to hold in practice, e.g. many real-world environments like robotics can be considered smooth~\citep{pirotta2015policy} and the same can be said about neural networks with ReLU activations~\citep{DBLP:journals/nn/PetersenV18}. Moreover, although the convexity assumption is much more restrictive, convexity is commonly assumed in RL~\citep{pinto2017robust,perolat2015approximate,patek1997stochastic} with minimax objectives and good empirical performance is still observed even when this assumption is not met.
Under Assumption~\ref{ass:smooth_convex}, we have the following lemma.
\begin{lemma}
\label{lemm:convex_f}
The expected return $\mathbb{E}[\mathcal{J}^\eta(\pi, \nu)]$ is convex with respect to $\pi$ and concave with respect to $\nu$.
\end{lemma}
\begin{proof}
This follows from the convexity of $\mathcal{R}$ and all $\mathcal{P}$.
\end{proof}

Building upon Lemma~\ref{lemm:convex_f}, we have the following result.

\begin{theorem}
\label{thm:convergence}
The proposed Stackelberg algorithm will converge towards the Minimax Equilibrium of the game.
\end{theorem}
\begin{proof}
This is a well-known result for Stackelberg games in the max-min setting with a convex-concave playoff function (see~\citet{fiez2019convergence}), a condition guaranteed by Lemma~\ref{lemm:convex_f}.
\end{proof}

Remember that the Minimax Equilibrium of the game contains a CVaR optimal policy per Theorem~\ref{thm:approx_equilibrium_cvar}.
Theorem~\ref{thm:convergence} therefore proves that the proposed gradient-based algorithm under the ACReL game formulation constitutes a theoretically justified way to attain a CVaR optimal policy.
A last theoretical result follows from the convex-concave objective in ACReL.

\begin{corollary}
For the proposed game, we have
\begin{equation*}
\min_{\nu} \max_\pi \mathbb{E} \left[\mathcal{J}^\eta(\pi, \nu)\right] =  \max_\pi \min_{\nu} \mathbb{E} \left[\mathcal{J}^\eta(\pi, \nu)\right].
\end{equation*}
\end{corollary}
\begin{proof}
This follows from the convexity Lemma~\ref{lemm:convex_f}, which allows to apply the minimax theorem~\citep{sion1958general}, inverting the $\min$ and $\max$ terms.
\end{proof}
This last corollary is of separate interest, as it notably implies that the Stackelberg algorithm can take any of the players as its leader to achieve the Minimax Equilibrium of the game.

\section{Experiments}
\label{sec:experiments}

We now demonstrate the applicability of ACReL with neural networks on an illustrative toy environment.
Specifically, we study the ability of an agent learning under ACReL to recover the true CVaR optimal policy and investigate the adversarial dynamics on the learning procedure.

\subsection{Experimental details}

\begin{figure}[h]
    \centering
    \includegraphics[width=0.3\textwidth]{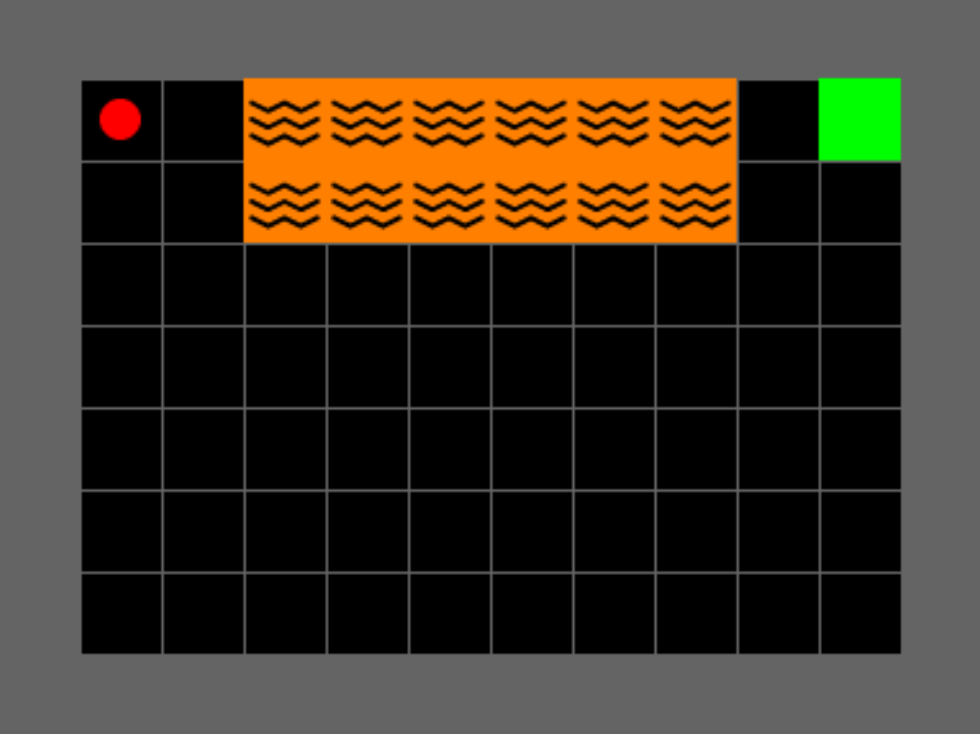}
    \caption{Overview of our Lava gridworld environment.}
    \label{fig:ACREL_Grid}
\end{figure}

We conduct experiments on a variation of the Gym Minigrid Lava environment~\citep{gym_minigrid}. Figure~\ref{fig:ACREL_Grid} illustrates this environment where an agent walks in a grid with the aim to reach the green goal tile as fast as possible from the red dot initial location, while avoiding stepping in the orange lava obstacle. 
An agent state corresponds to the $(x, y)$ coordinates of the agent in the grid. 
The action space for the agent corresponds to moving into one of the four cardinal directions. 
At each time step $t$ of a trajectory (episode), the agent incurs a reward penalty of $-0.035$, motivating the agent to reach the goal using the shortest path. The episode ends when the agent either reaches the goal, generating a reward of $+1$, falls into the lava, generating a reward of $-1$, or when the maximum number of $40$ steps is reached within the episode, without any additional reward.

\paragraph{Stochasticity}
The environment is stochastic such that the agent performs random action (instead of the action recommended by the policy $\pi$) with probability $p$.
This environment poses an intrinsic risk dilemma to the agent. Indeed, as the agent incurs penalties for every step taken, they want to walk the shortest path, which is close to the lava zone. However, the intrinsic environment stochasticity ($p$) may push the agent into the lava.
This dilemma between minimizing the risks of falling into the lava while also minimizing the number steps for reaching the goal serves as a good minimal illustration of a safety-critical environment.
The more importance given by the agent to avoid large negative rewards, the more willing the agent will be to avoid the cliff by walking a longer patch (towards the bottom of the grid).
We hereby report the results of experiments with $p=0.1$
Further results using $p=0.05$ can be found in Appendix~\ref{app:add_res}.

\paragraph{Risk-sensitivity}
We consider three different confidence levels: $\alpha \in \{ 1, 0.04, 0.01 \}$.
The CVaR$_1$ objective corresponds to the classical risk-neutral RL objective whereas CVaR$_0.04$ and CVaR$_0.01$ are highly risk-sensitive objectives.
In ACReL, these objectives respectively translate into initial adversarial budgets $\eta=1$, $\eta=25$, and $\eta=100$.

\paragraph{ACReL}
Both the agent and the adversary are trained jointly following Algorithm~\ref{alg:algorithme} with $K_\text{adv}=2$, according to a cross-validation over $K_\text{adv} \in \{2,4,8\}$. Both policies are learned using Actor-Critic PPO~\citep{schulman2017proximal}.

\paragraph{Baselines}
We employ IQN-CVaR~\citep{dabney2018implicit} as a first baseline, the CVaR variant of a state-of-the-art risk sensitive distributional RL algorithm.
Since our goal is to determine the optimality of the learned policies with respect to the CVaR objective, we also compute the true optimal policies via policy iteration~\citep{chow2015risk}.

To ensure fair comparison, both ACReL and IQN-CVaR are trained under a similar training regime.
Namely, all algorithms are executed for a maximum of 5m steps in the environment and agents are all implemented as a single-layer neural network with 64 hidden neurons. 
In order to assess the robustness of the approaches, we run each algorithm using $n=10$ random seeds.
Our codebase is attached as supplementary material for full reproduction of the experiments and figures.
Further details on algorithms and training hyperparameters are discussed in Appendix~\ref{app:exp_details}.




\subsection{Learned policies}

\begin{figure*}[ht!]
\centering
\begin{subfigure}[t]{0.3 \linewidth }
    \includegraphics[width=\textwidth]{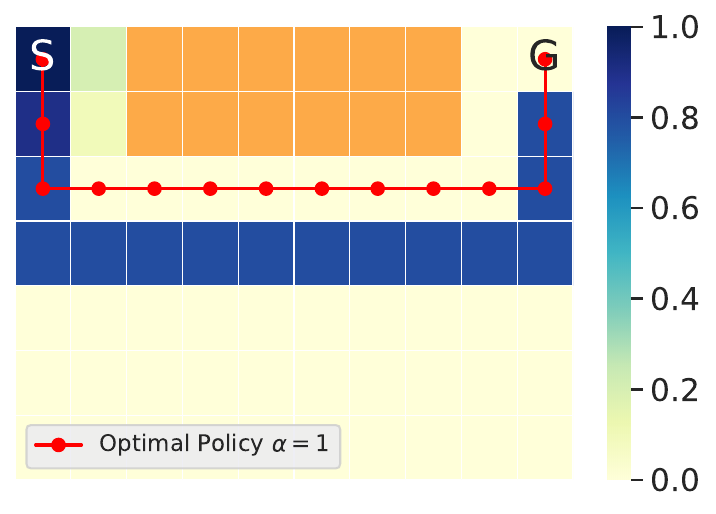}
    \caption{ACReL$_1$ ($\eta=1$)}
    \includegraphics[width=\textwidth]{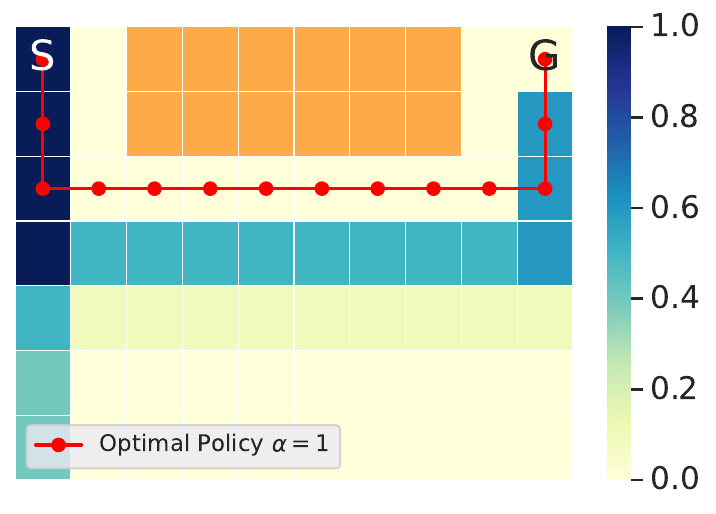}
    \caption{IQN-CVaR$_{1}$}
\end{subfigure}
\hspace{0.03 \linewidth}
\begin{subfigure}[t]{0.3 \linewidth}
    \includegraphics[width=\textwidth]{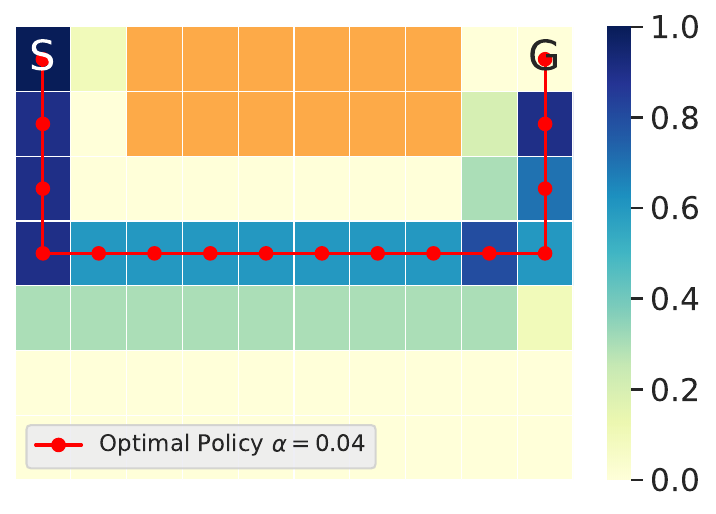}
    \caption{ACReL$_{0.04}$ ($\eta=25$)}
    \includegraphics[width=\textwidth]{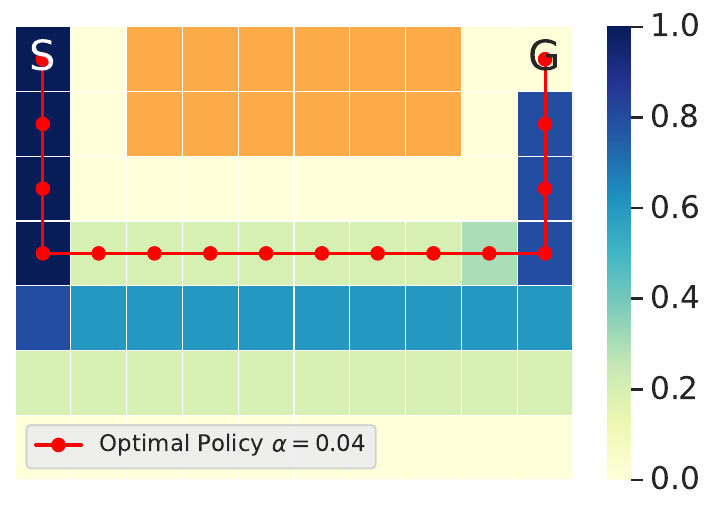}
    \caption{IQN-CVaR$_{0.04}$}
\end{subfigure}
\hspace{0.03 \linewidth}
\begin{subfigure}[t]{0.3 \linewidth}
    \includegraphics[width=\textwidth]{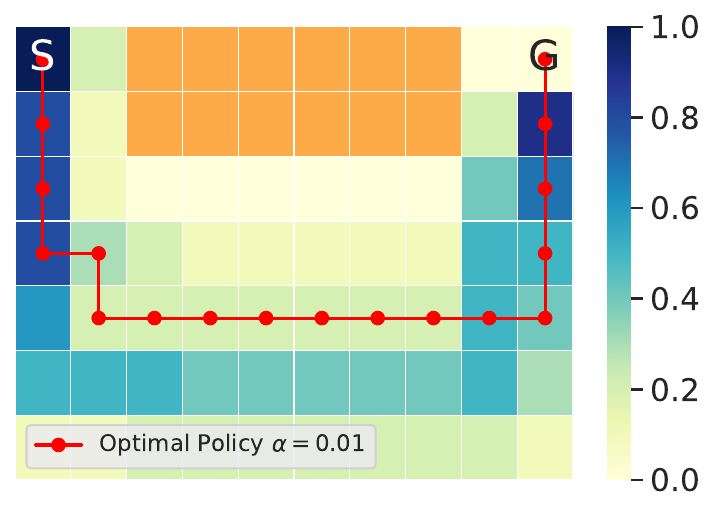}
    \caption{ACReL$_{0.01}$ ($\eta=100$)}
    \includegraphics[width=\textwidth]{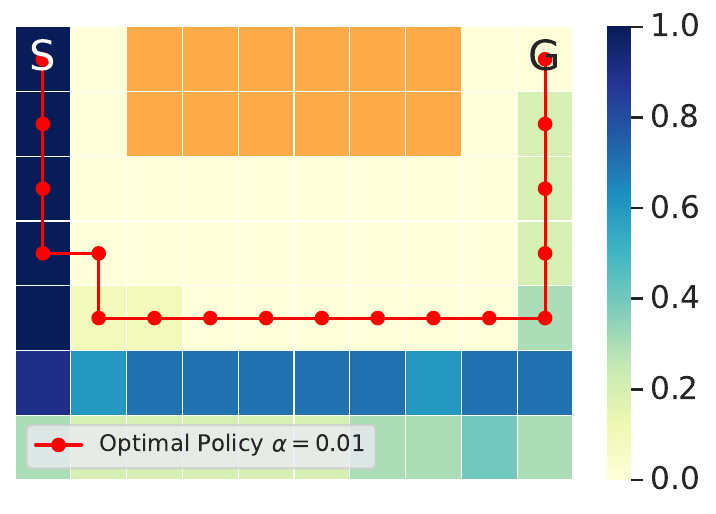}
    \caption{IQN-CVaR$_{0.01}$}
\end{subfigure}
\caption{Agent policies learned with ACReL (a, c, e) and IQN-CVaR (b, d, f) for different CVaR$_\alpha$ objectives with $p=0.1$.
}
\label{fig:averaged_seeds}
\end{figure*}

We visualize the agent policies learned after 5m steps by executing each policy in a deterministic environment ($p=0$) without any adversary. 
This allows to observe the decisions made by the agent without the noise induced by environment stochasticity.
Figure~\ref{fig:averaged_seeds} shows the resulting policies learned with $p=0.1$ (see Figure~\ref{fig:averaged_seeds005} in Appendix~\ref{app:add_res} for $p=0.05$) using blue shading to indicate the distribution of grid locations visited by the policies, averaged over the random seeds, and red paths indicating the CVaR$_{\alpha}$ optimal policies.

As expected, we observe that the optimal policies adopt an increasingly cautious behavior as $\alpha$ decreases,  the paths progressively moving further away from the lava.
First thing to note is that this tendency is also found both with ACReL and IQN-CVaR.
For the CVaR$_1$ and CVaR$_{0.01}$ objectives, both algorithms learn policies that are slightly over-cautious.
Interestingly, for CVaR$_{0.04}$, ACReL converges more closely to the true optimal policy than IQN-CVaR.
This confirms that the performance of ACReL in retrieving CVaR$_\alpha$ optimal policies in practice matches the current state-of-the-art in the literature. 

However, further investigation reveals that both ACReL and IQN-CVaR suffer from convergence instabilities.
More specifically, for both algorithms, some runs result into sub-optimal policies even after 5m time steps.
We believe that those issues arise from the inherent difficulty of the task, which strongly depends on exploration due to the simplistic agent state representation.
Indeed, since the agent only observes its current position, the goal can only be reached ``by chance'' the first time. Therefore, one unlucky agent may seldom observe the goal state reward and learn a policy that does not aim for the goal state, but rather simply tries to avoid getting pushed in the lava.
Despite these challenges, both strategies converge in the vast majority of runs (see Appendix~\ref{app:add_res}).



\subsection{Adversarial dynamics} 

\begin{figure}[h!]
\centering
\begin{subfigure}[t]{0.2 \textwidth}
    \makebox[20pt]{\raisebox{35pt}{\rotatebox[origin=c]{90}{With adversary}}}%
    \includegraphics[width=\textwidth]{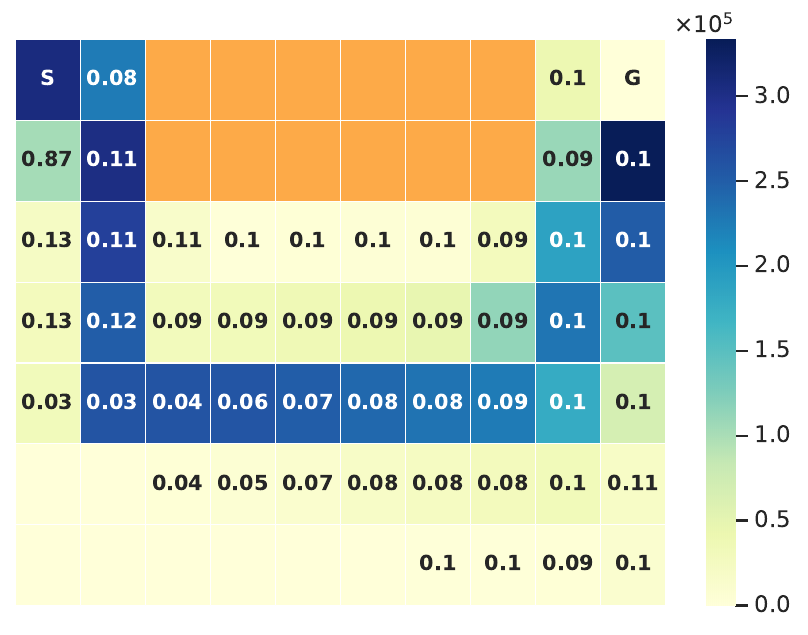}
    \makebox[20pt]{\raisebox{35pt}{\rotatebox[origin=c]{90}{No adversary}}}%
    \includegraphics[width=\textwidth]
    {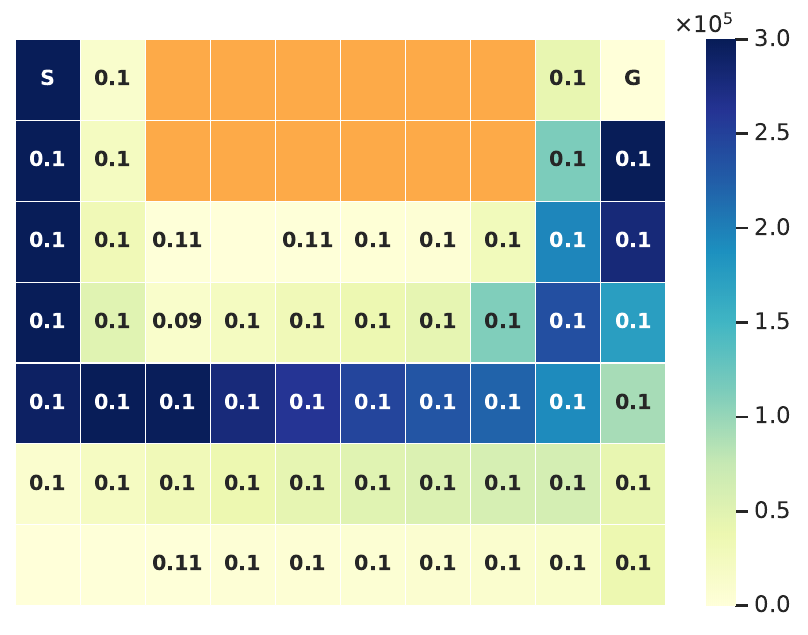}
    \caption{ACReL$_{0.04}$ ($\eta=25$)}
\end{subfigure}
\hspace{3em}
\begin{subfigure}[t]{0.2 \textwidth}
    \includegraphics[width=\textwidth]{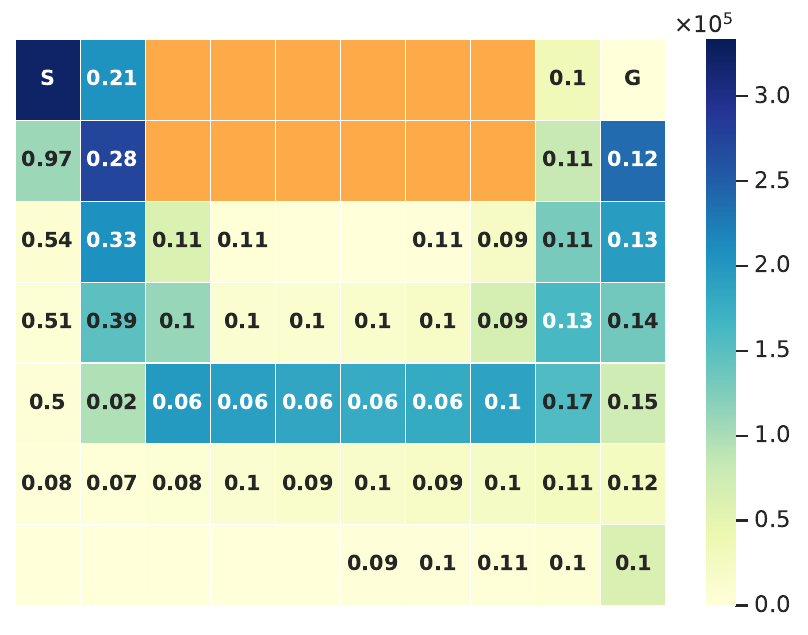}
    \includegraphics[width=\textwidth]
    {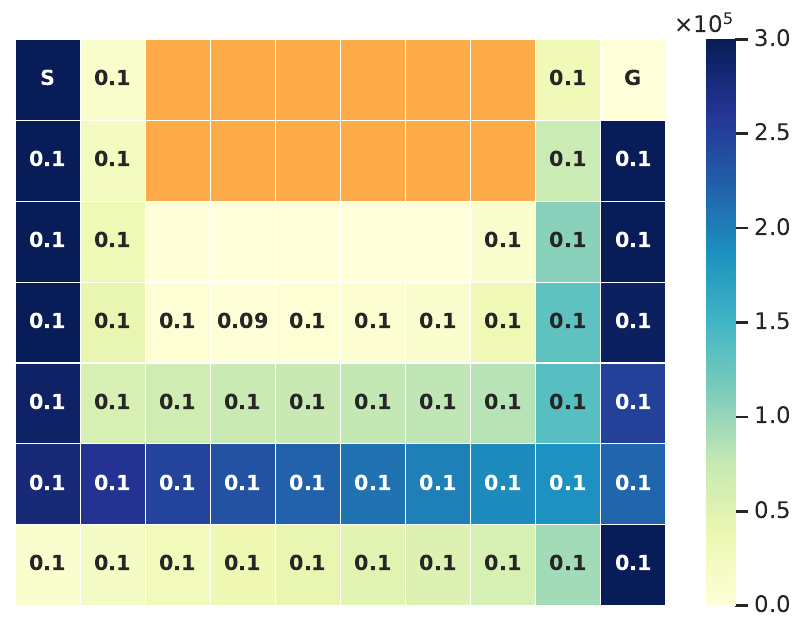}
    \caption{ACReL$_{0.01}$ ($\eta=100$)}
\end{subfigure}
\caption{
Agent policies learned with ACReL deployed on the stochastic environment ($p=0.1$), with and without the adversary.
}
\label{fig:empirical_view}
\end{figure}

Having established that ACReL is competitive when it comes to retrieving CVaR$_\alpha$ optimal policies, we now study the adversarial policy and its impact on the agent policy.
To achieve this we first visualize the ACReL agent policies learned after 5m steps by executing them in the stochastic environment to collect 300k trajectories, with and without the adversary.
Figure~\ref{fig:empirical_view} shows the resulting policies (learned with $p=0.1$) using blue shading to indicate the number of times each grid position was visited over all trajectories, averaged over the random seeds.
Numbers indicate the probability that the agent may perform an action different from the action recommended by the policy $\pi$.
In order to display accurate estimates, only the probabilities for the locations that the agent visited at least $1500$ times are displayed.
Probabilities greater than 0.1 (environment stochasticity $p$) indicates that the adversary is spending some of its budget to alter the agent policy in this location.

As expected, we first observe that the agent is pulled closer to the lava zone in the presence of an adversary (top row).
Note that the adversary is mostly active at the beginning of a game episode, regardless of the amount of perturbation budget. 
The beginning of the game is indeed the riskiest situation for the agent since the starting point is very close to the lava (top-left corner).
More sophisticated adversarial strategies emerge when the adversary benefits from a higher budget.
When the agent is far from both the lava and the goal, the perturbation probability drops below $0.1$ (the natural environment stochasticity), meaning that the adversary tends to give more leverage to the agent in this zone. 
This allows the adversary to recover some budget and become more powerful when the agent gets closer to the goal, hence closer to the lava.
This is also confirmed by looking closely at the adversarial policy over single trajectories (Appendix~\ref{detailed_analysis_adversarial_policy}).

\section{Related Work}
\label{sec:related_work}

\paragraph{Adversarial RL}
Although typically used in supervised learning, adversarial learning~\citep{kurakin2016adversarial} has been applied in RL under the robust MDP formulation.
In robust MDP the agent aims to optimize the worst-case return of a policy under some uncertainty limitations, implemented as bounded adversarial perturbations.
These perturbations can occur over state representations~\citep{pinto2017robust,kamalaruban2020robust}, the policy~\citep{mandlekar2017adversarially,tessler2019action,li2019robust}, or the reward function~\citep{lin2020model}.
However, since the robust MDP framework defines its perturbation limitation over every single time step, setting the actual limitation value is a challenging task.
Because the adversary may reach its perturbation limit at every step, a large limit may produce overly conservative policies while a small limit may hamper the actual robustness of the learned policy~\citep{mozian2020learning}.
In contrast, our proposed adversary budget is defined over a whole trajectory, ensuring that the worst possible adversarial perturbation cannot occur at every step, an arguably more plausible scenario.


\paragraph{CVaR RL}
Optimization of the CVaR criterion in RL have been extensively studied in the tabular and linear settings.
The proposed algorithms range from policy gradient~\citep{tamar2015policy} and $Q$-learning~\citep{chow2015risk}, to actor-critic methods~\citep{tamar2013variance}.
While these approaches are all limited to either tabular or low-dimensional action and state space settings, the gradient-based algorithm considered in this work is compatible with neural network settings. 
Interestingly, \cite{chow2015risk} proved a link between a policy's robustness to modelling errors of varying strength and CVaR optimality.
Although our adversarial model is inspired by this result, our proposed game is distinguished by the key fact that our adversary is a learnable player.

More recently, distributional RL~\citep{bellemare2017distributional,dabney2018implicit} has been used to learn CVaR optimal policies in high-dimensional settings~\citep{keramati2020being,zhang2021safe}.
This approach has seen growing usage, largely due to its empirical efficiency and its ability to incorporate a wide variety of risk measures in its objective, not exclusively the CVaR.
However, since distributional RL does not generally converge over the whole return distribution~\citep{bellemare2017distributional}, any risk measure optimization on the learned distribution is not theoretically guaranteed to converge to the true optimally risk-averse policy~\citep{dabney2018implicit}. In this work, we provide a theoretical guarantees on the optimality of the policy obtained from the proposed strategy.

The closest work to ours would be Robust Adversarial Reinforcement Learning (RARL)~\citep{pinto2017robust}, where a policy is learned by allowing an adversary to apply external forces to the environment according to a CVaR objective.
The action space of the adversary is therefore different from the one in ACReL, as it applies various problem-specific perturbations to the model.
This adversarial formulation notably implies that their CVaR objective is only an informal one, as one cannot link the perturbation range given to the adversary to a specific CVaR$_\alpha$ objective.
In contrast, we establish a clear, theoretically justified, connection between the budget of the adversary and the $\alpha$-quantile optimized by the agent.

\section{Conclusion}
\label{sec:conclusion}

In this work, we introduced ACReL, an adversarial game designed to compute risk-sensitive policies with respect to the Conditional Value-at-Risk (CVaR) risk. 
This adversarial game limits an adversary's perturbations to be within a specified budget over entire trajectories and to only change next state transitions for an agent.
We proved that the the resulting policy at the game's equilibrium is CVaR optimal, with the result also holding around the neighborhood of the equilibrium.  
We then showed a gradient-based algorithm using the Stackelberg formulation of the game to solve the game, both proving its convergence under sufficient conditions and presenting key considerations for a practical implementation.
We put our method to the test on an artificial risky environment, illustrating that ACReL matches the performance of a state-of-the-art baseline in successfully retrieving CVaR optimal policies.

Given the fact that this is the first game-theoretic perspective on CVaR RL, we expect this paper to be a building block towards connecting risk-sensitivity in RL with the field of Game Theory.
Notably, we believe it may be possible to link our adversarial framework with other risk-sensitivity measures by allowing the budget to vary during training instead of keeping it fixed. 

One possibly interesting application of our method is with regards to the Sim2Real domain, where one wishes to optimize to train an RL agent on a high-quality simulation environment before real-life deployment.
Since this domain is naturally concerned with risk-sensitivity and supposes access to a simulator to access and perturbate the true next state transitions, it appears highly likely that our proposed approach can be of use in this context.
In future work, we would like to apply ACReL to the Sim2Real application to study the scalability of ACReL, both with respect to task difficulty and neural networks complexity.

\clearpage
\bibliography{acrel_uai22}

\clearpage
\appendix

\section{Practical Algorithm Details}
\label{app:prac_algo_details}

\paragraph{Practical Stackelberg Algorithm}
Recall the Stackelberg optimization problems for the leader
\begin{equation*}
\theta_L \in \argmax_\theta \left\{ u_L(\theta, \theta_F^{\star}) \, \, \Bigg | \, \, \theta_F^{\star} \in \argmax_{\theta_F} u_F(\theta, \theta_F) \right\}
\end{equation*}
and for the follower
\begin{equation*}
\theta_F \in \argmax_{\theta} u_F(\theta_L, \theta).
\end{equation*}

As is usually done in gradient-based methods, we may solve the leader's optimization problem using gradient ascent.
Making sure that the follower remains optimal in between leader updates, we use the following bi-level procedure:
\begin{align}
    & \theta_F^{k+1} \approx \argmax_\theta u_F(\theta_L^k, \theta)\label{eq:fol_pro}\\
    & \theta_L^{k+1} = \theta_L^k + \beta \frac{\partial}{\partial \theta} u_L(\theta, \theta_F^{k+1})\label{eq:lead_pro}
\end{align}

where $\beta$ is a learning rate hyperparameter, $\frac{\partial }{\partial \theta}$ is a partial derivative gradient and updates are done sequentially for each parameters.
Note that since the leader update \ref{eq:lead_pro} is done using $\theta_F^{k+1}$, optimal with respect to $\theta_L^k$ by definition, alternating between these two updates does indeed respect the Stackelberg game requirements.

Applying the above bi-level procedure remains challenging in practice however.
First, the $\argmax$ operation in \ref{eq:fol_pro} is actually an optimization problem on its own which may require an impractical amount of time to compute.
Based on the intuition that best responses for consecutive leader parameters should be near each other for a small enough $\beta$, we approximate the $\argmax$ by doing a few gradient ascent steps.
Second, while the leader's partial gradient $\frac{\partial }{\partial \theta} u_L(\theta, \theta_F^{k+1})$ may be computed exactly to incorporate the implicit relation between the leader and follower, this exact update requires the computation of a Jacobian matrix which is computationally intensive.
Instead, we simply use a a first-order estimate, using the total gradient $\nabla_\theta u_L(\theta, \theta_F^{k+1})$, which is biased may be computed more efficiently.

\paragraph{Perturbation rescaling and budget constraint}
At every time step $t$, the adversary produces perturbations $\delta_t = \nu(\mathcal{P}_t, \eta_t)$.
Recall that, in order to be admissible, the adversary's output $\delta_t$ has to produce perturbed transitions $\hat{\mathcal{P}}_t = \mathcal{P} \circ \delta_t$ that are a valid distribution and every $\delta_i$ has to be less than the remaining budget $\eta_t$.

In order to respect both constraints simultaneously, we propose a two-step normalization procedure to ensure the final output is an admissible perturbation.
We begin by normalizing the adversary's raw output $y$ to yield a perturbation $\delta$ that makes $\hat{\mathcal{\mathcal{P}}} = \mathcal{P} \circ \delta$ a distribution\footnote{Note that this chosen normalization requires $y_i > 0$ for all $i$ and may be achieved through an exponential output activation for instance.} $\delta_i = y_i/\sum_j \mathcal{P}_j y_j$.

If the $\delta$ generated by the normalization respects the budget constraint, then we can return $\delta$ directly. However, if the perturbation $\delta$ is over budget, we have to reduce it to a version $\tilde{\delta}$ within budget.
Exploiting the convexity of probability spaces, we apply the following transformation:
\begin{equation}
    \tilde{\delta}_i = \frac{\delta^\star - \eta_t + (\eta_t - 1)\delta_i}{\delta^\star - 1},
\label{eq:corrected_perturbations}
\end{equation}
where $\delta^\star := \max_i \delta_i$.
The transformation \ref{eq:corrected_perturbations} benefits from the following properties:
\begin{enumerate}
    \item it does not move the unperturbed transitions ($\tilde{\delta}_i=1 \iff \delta_i =1$);
    \item it is strictly increasing (i.e. $\tilde{\delta}_i > \tilde{\delta}_j \iff \delta_i > \delta_j$);
    \item it will bring each $\delta_i$ closer to 1 (i.e. $|\delta_i - 1| > |\tilde{\delta}_i - 1|$).
\end{enumerate}
The obtained properties ensure that the overall intuition behind the adversary's proposed perturbation is kept when applying the correction.

\section{Experimental details}
\label{app:exp_details}

\begin{figure}[ht]
    \centering
    \includegraphics[width=0.3\textwidth]{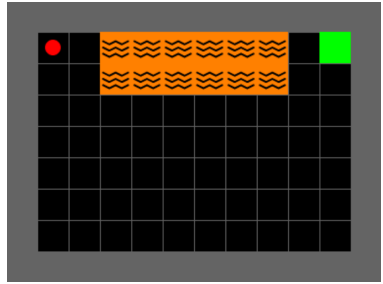}
    \caption{Overview of our Lava gridworld environment.}
    \label{fig:CARL_grid_app}
\end{figure}

\paragraph{Toy Risky Environment}
Figure \ref{fig:CARL_grid_app} is a visual representation of the Lavagrid environment used to conduct our experiments.
The agent's starting position, referred to as \textbf{S} in the main paper, is shown here as a red dot.
The goal tile where the agent receives a large reward, referred to as \textbf{G} in the main paper, is shown here as a green square.
Wavy orange tiles represent "lava" squares, which the agent intuitively learns to keep away from because they produce a large negative reward.

In our experiments, the agent takes as input their current coordinates $(x,y)$ in the grid which is a vector of length 2 and returns an action as a one-hot-vector of size 4.
For ACReL, the adversary takes as input a concatenation of $s_t$, the value of their remaining perturbation budget $\eta_{t}$, and the probabilities $p_t$ that each of the 4 possible actions will be the one executed, making the adversary input a vector of 7 elements. 
The adversary returns perturbed probabilities $\hat{p}_t$ over possible action executions.
This particular choice of representation for the agent's state and action spaces were made after initial experiments proved unsuccessful when using the complete next state distributions $\mathcal{P}_t$.
We estimate that sparse transitions like the ones encountered in our selected gridworld make the adversary task more difficult.
This is why we proposed the state and action spaces described above for the adversary, which have the benefit of being drastically smaller while preserving all the necessary state information and action range.

In order to evaluate the robustness of both methods, we replicate each instance of the game 10 times, each time with a different random seed. The environment is stochastic with probability ($p = 0.1$) for the agent to perform a random action instead of the chosen action $a_t$. For ACReL, the environment's seed (governing its randomness), the agent (governing the initialization of its policy and the action selection process), and for adversary's (governing the initialization of its policy and the perturbation prediction process) are reinitialized. For IQN-CVaR, the environment's seed (governing its randomness), as well as the agent's (governing the initialization of its policy) are reinitialized.

\paragraph{ACReL}
Both the agent and the adversary are trained using PPO~\citep{schulman2017proximal} with shared hyperparameters. 
For both players, the actor and critic are Multi-Layer Perceptrons (MLP) with 1 hidden layer of 64 nodes and \textit{tanh} as activation function. 
A clipping value of $0.2$ is used. 
The Generalized Advantage Estimator hyperparameter $\lambda$ is set to $0.95$. 
The entropy term coefficient is set to $0.1$ and the value loss term coefficient to $0.5$. 
The maximum norm of the gradient is set to $0.5$. 
Both players use an Adam optimizer with their own decaying learning rate that decreases linearly from $0.005$ to $0.0001$ over the course of training. 
The $\varepsilon$ for Adam is set to $1e-8$. 

During training, an update of the parameters occurs every 10k time steps. 
The adversary is updated  $K_{adv}=2$ times in between each agent's update (see in Algorithm \ref{alg:algorithme}). 
For the update, the batch size is set to 256 observations and there are 4 gradient steps done.

\paragraph{IQN-CVaR}

We use a publicly available library, Tianshou~\citep{weng2021tianshou}, for the implementation of IQN,
We simply make the modifications on the provided implementation to match the IQN-CVaR algorithm as described in the original paper~\citep{dabney2018implicit}.
The network architecture is the same as in ACReL also using an Adam optimizer.
The training procedure is separated in two portions.
In the first portion, lasting for the first 2.5M steps, the $\varepsilon$ value in the $\varepsilon$-greedy action selection linearly decreases from $1.0$ to $0.1$ while the learning rate is set at $0.002$.
In the second portion, lasting for the last 2.5M steps, $\varepsilon$ is kept fixed at $0.1$ and the learning rate is lowered at $0.0002$.

Updates are done every 10 time steps by sampling a batch of $B=128$ transitions from a replay buffer with a maximal capacity of $500~000$ transitions.
For all other hyperparameters not named (e.g. the number of $\tau$ quantile samples), we keep Tianshou's default values.

\paragraph{Compute requirements}
All experiments are run on a single computer equipped with a Nvidia P100 GPU, using 4 CPU cores and less than 5GB of RAM.
Each training run takes about 3h for ACReL and 1h for IQN-CVaR.
It is to be expected that ACReL takes slightly more time to train, for the simple reason that each time step in the environment requires calling two neural networks (the agent and adversary) instead of a single one with IQN.
The policy visualization experiments ran using 300k trajectories for ACReL take around 3h to complete as well.

\section{Additional Results}
\label{app:add_res}

\subsection{Detailed analysis of the adversarial policy}
\label{detailed_analysis_adversarial_policy}

\begin{figure*}[h!]
\centering
\begin{subfigure}[b]{0.29\linewidth}
 \centering
 \includegraphics[width=\textwidth]{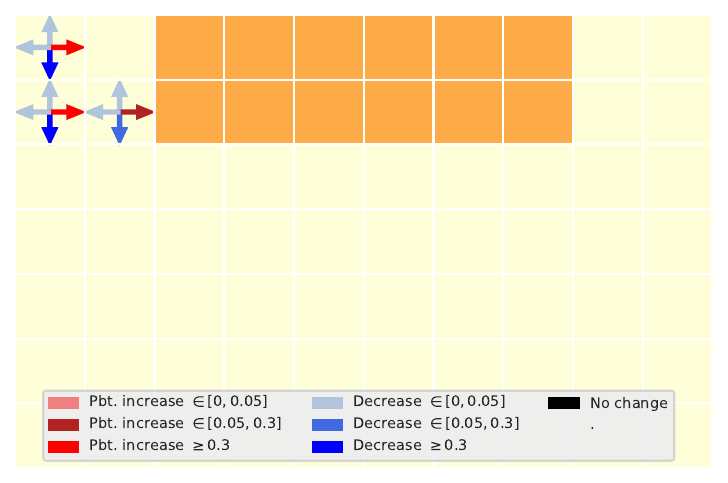}
 \caption{Adversary wins}
\end{subfigure}
\begin{subfigure}[b]{0.29\linewidth}
 \centering
 \includegraphics[width=\textwidth]{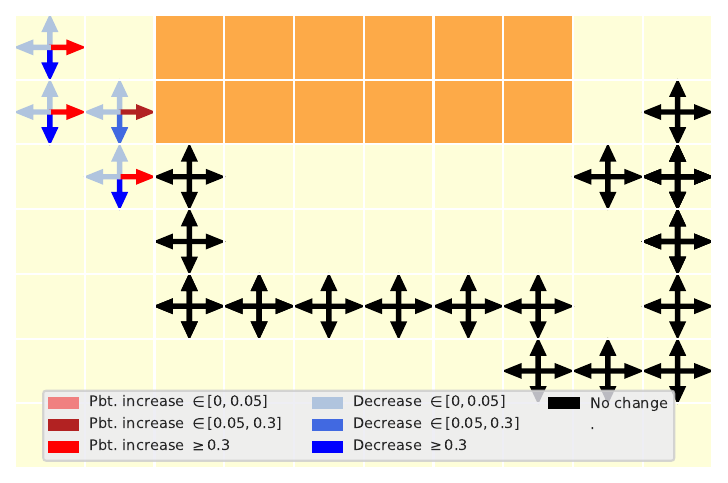}
 \caption{No budget left}
\end{subfigure}
\begin{subfigure}[b]{0.29\linewidth}
 \centering
 \includegraphics[width=\textwidth]{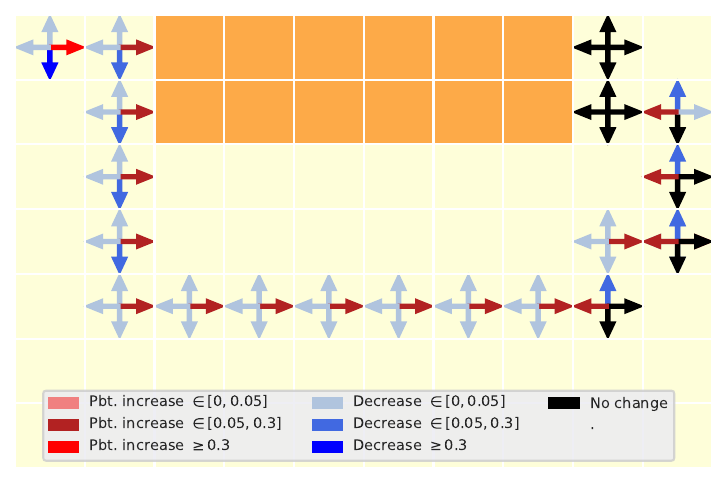}
 \caption{No budget left}
\end{subfigure}
\caption{Example interactions between the adversary and agent in ACReL. Colored arrows illustrate the amplitude of perturbation applied to the initial probabilities. 
}
\label{fig:cherry_picked100}
\end{figure*}

For a finer-grain analysis of the adversarial dynamics, we detail in Figure~\ref{fig:cherry_picked100} interesting interactions between the agent and adversary over a single trajectory.
We notice three distinct interaction patterns.
The first pattern corresponds to the situation where the adversary spends all their budget at the beginning of the episode and successfully wins the game by pushing the agent into the lava. 
The opposite interaction occurs when the adversary spends all their budget at the beginning of the episode but fails to make the agent hit lava. 
In this case, since the adversary cannot influence the agent anymore, the trajectory continues as the agent pleases, with the usual intrinsic environment stochasticity. 
A third and perhaps most interesting interaction occurs when the adversary spends a large amount of budget at the beginning but keeps enough budget to remains able to act during the rest of the episode.
Interestingly, the adversary increases the winning direction probability (right) which apparently helps the agent reaching their goal. 
In the end of the episode, when the agent has to move closer to the lava before reaching the goal state, the adversary plays a final move, spending all its budget towards moving the agent directly next to the lava.

\subsection{In-Depth Run Analysis}
\label{more_convergence_comparaison}

\begin{figure*}
\centering
  \begin{subfigure}[c]{0.3\textwidth}\centering
    \myrowlabel{$S = 1$}
    \raisebox{-.5\height}{\includegraphics[width=.5\textwidth]{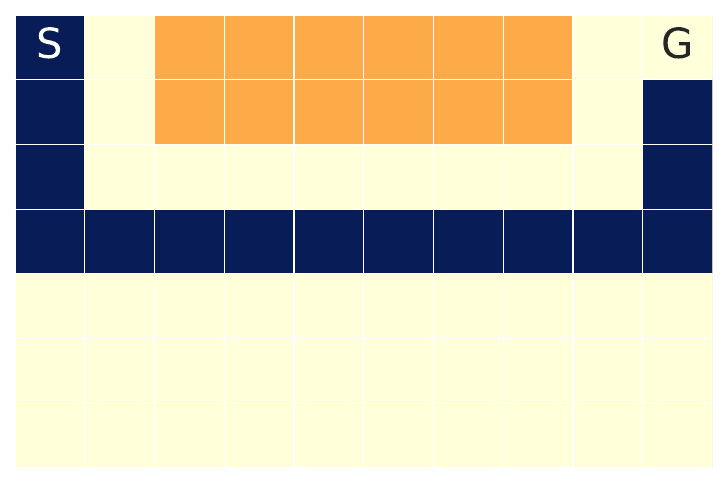}}\\
    \myrowlabel{$S = 2$}
    \raisebox{-.5\height}{\includegraphics[width=.5\textwidth]{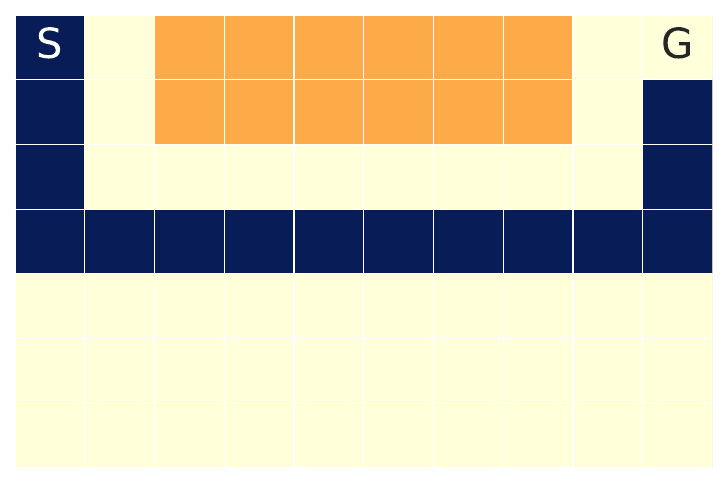}}\\
    \myrowlabel{$S = 3$}
    \raisebox{-.5\height}{\includegraphics[width=.5\textwidth]{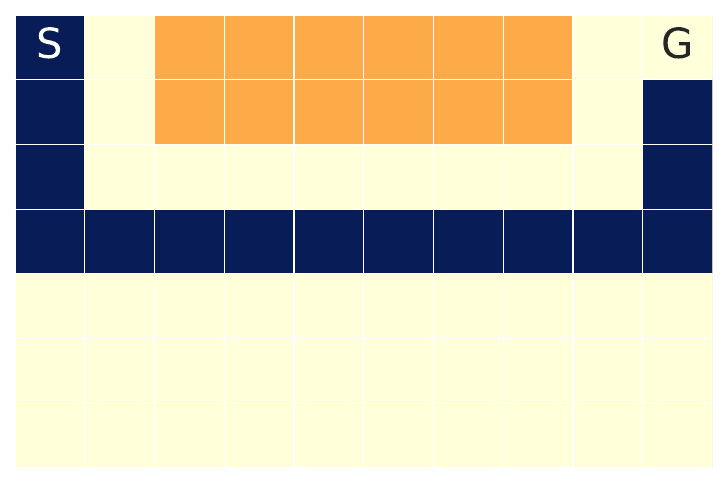}}\\
    \myrowlabel{$S = 4$}
    \raisebox{-.5\height}{\includegraphics[width=.5\textwidth]{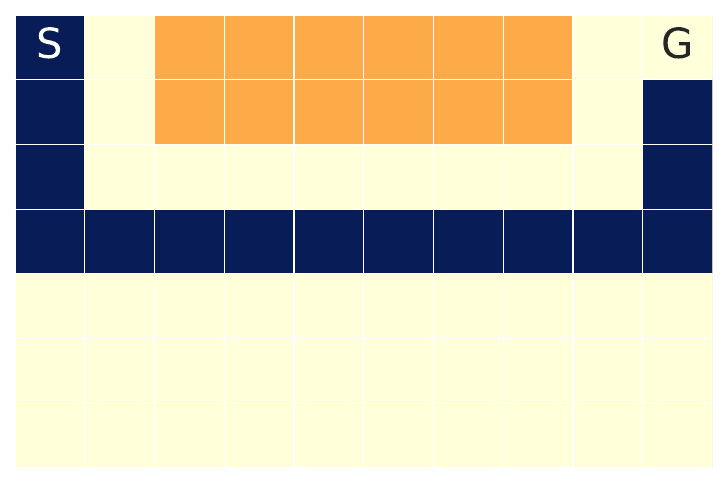}}\\
    \myrowlabel{$S = 5$}
    \raisebox{-.5\height}{\includegraphics[width=.5\textwidth]{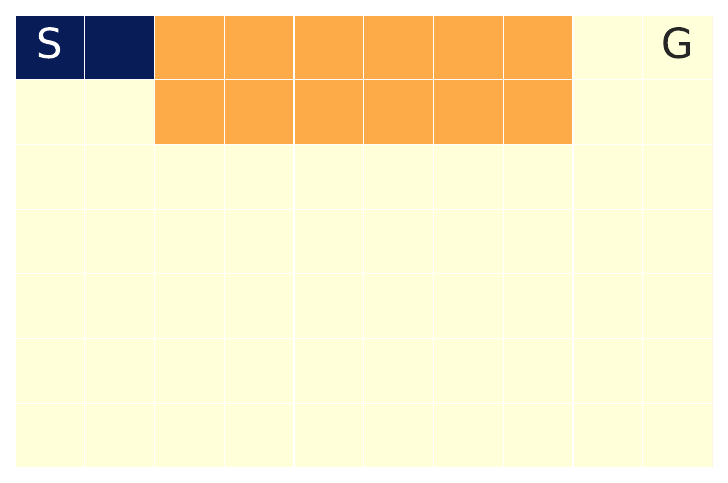}}\\
    \myrowlabel{$S = 6$}
    \raisebox{-.5\height}{\includegraphics[width=.5\textwidth]{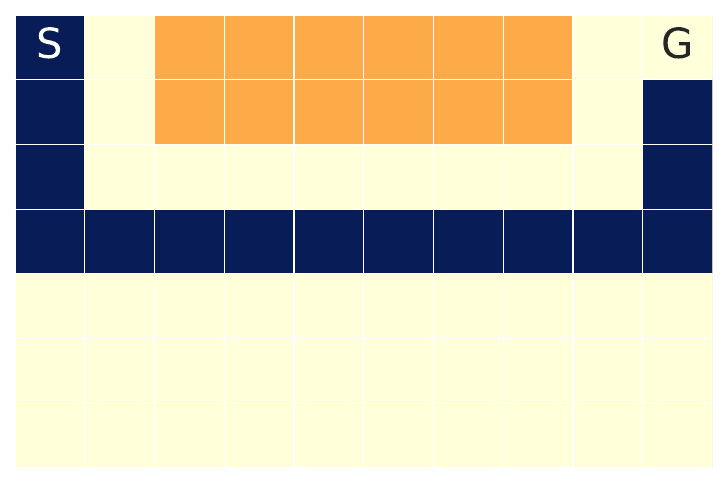}}\\
    \myrowlabel{$S = 7$}
    \raisebox{-.5\height}{\includegraphics[width=.5\textwidth]{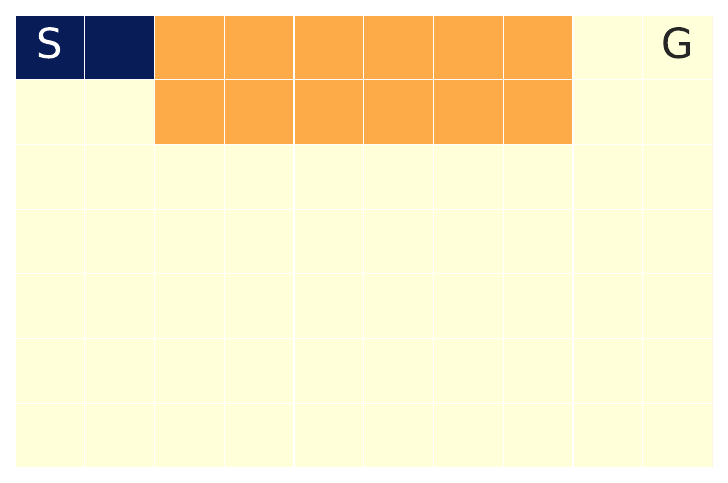}}\\
    \myrowlabel{$S = 8$}
    \raisebox{-.5\height}{\includegraphics[width=.5\textwidth]{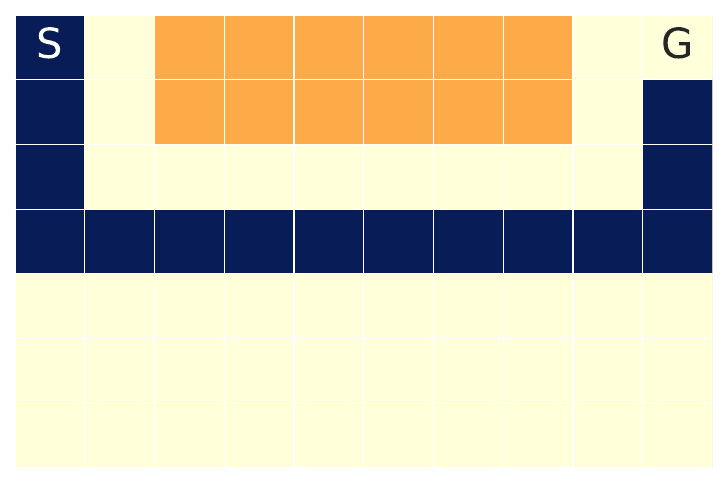}}\\
    \myrowlabel{$S = 9$}
    \raisebox{-.5\height}{\includegraphics[width=.5\textwidth]{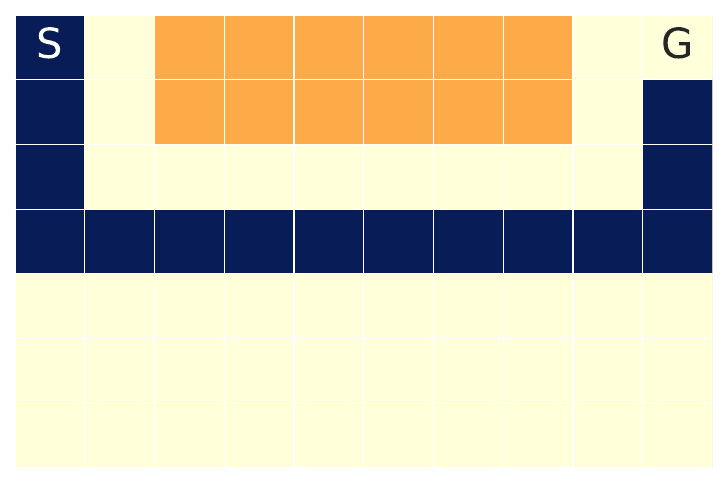}}\\
    \myrowlabel{$S = 10$}
    \raisebox{-.5\height}{\includegraphics[width=.5\textwidth]{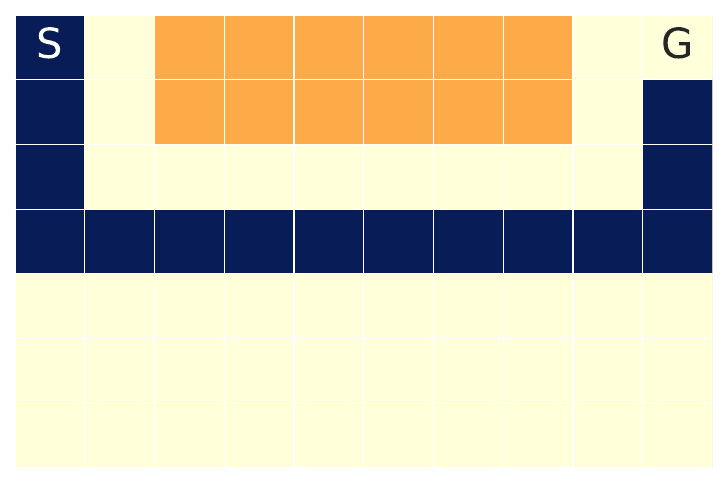}}\\
    \caption{ACReL$_1$ ($\eta = 1$)}
\end{subfigure}%
\hspace{-2em}
\begin{subfigure}[c]{0.3\textwidth}\centering
    \includegraphics[width=.5\textwidth]{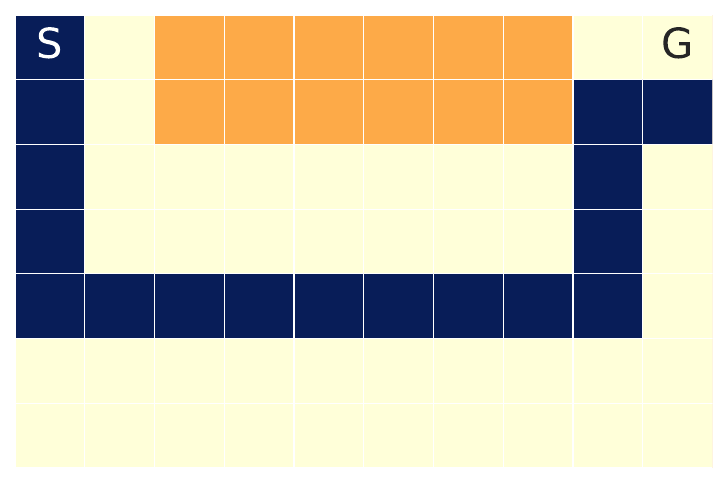}
    \includegraphics[width=.5\textwidth]{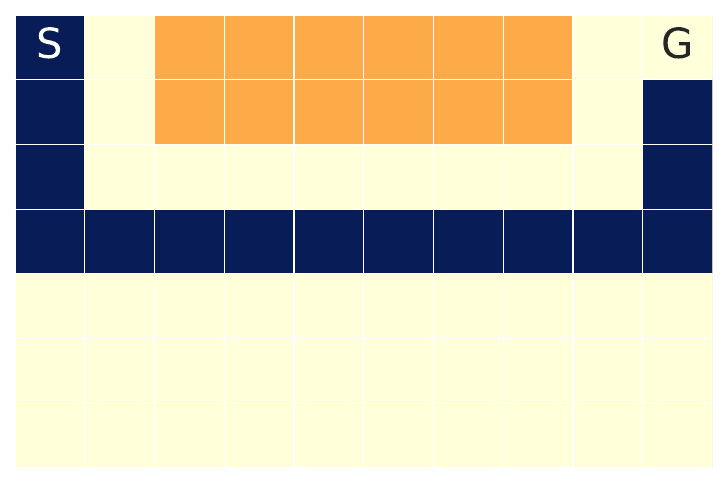}
    \includegraphics[width=.5\textwidth]{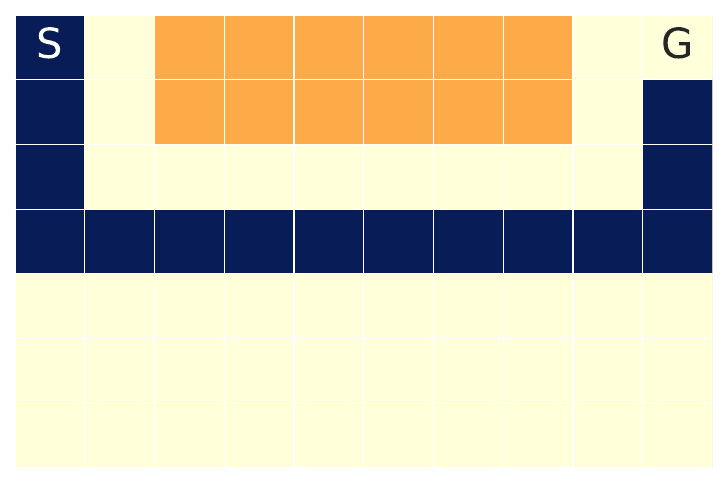}
    \includegraphics[width=.5\textwidth]{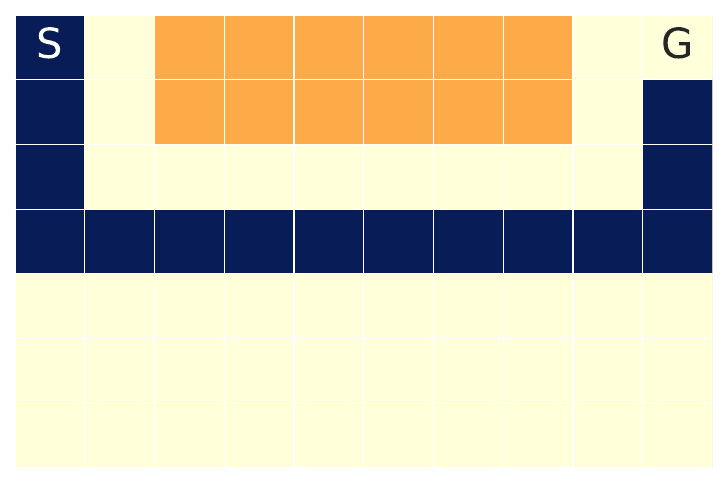}
    \includegraphics[width=.5\textwidth]{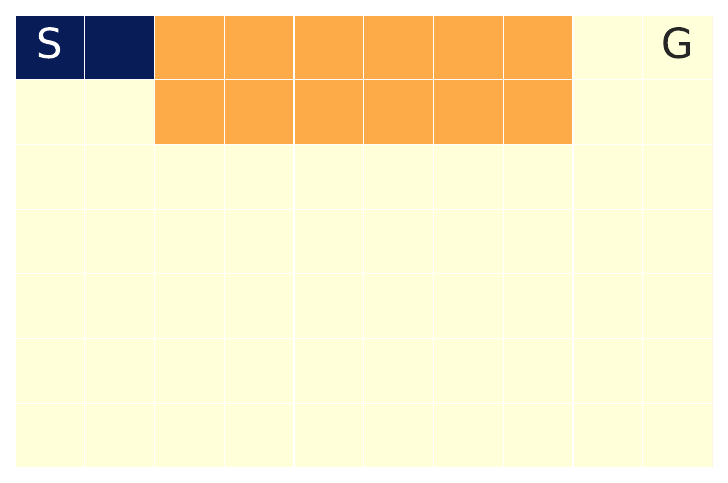}
    \includegraphics[width=.5\textwidth]{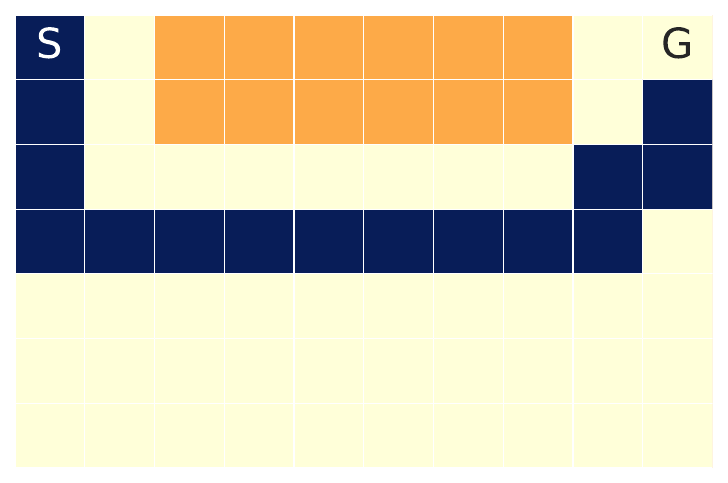}
    \includegraphics[width=.5\textwidth]{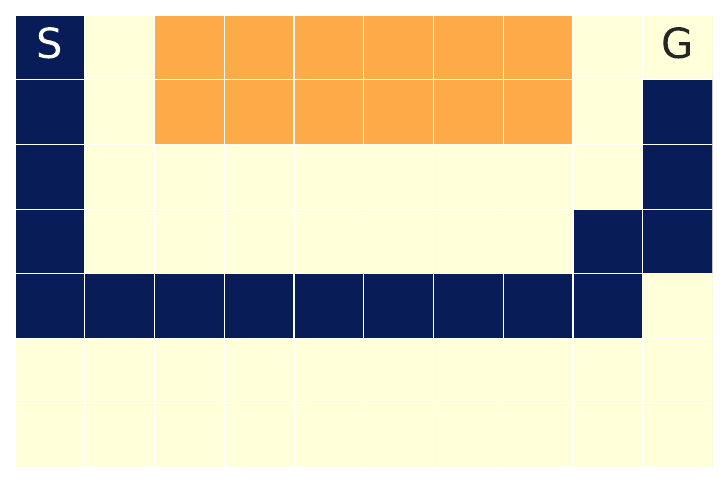}
    \includegraphics[width=.5\textwidth]{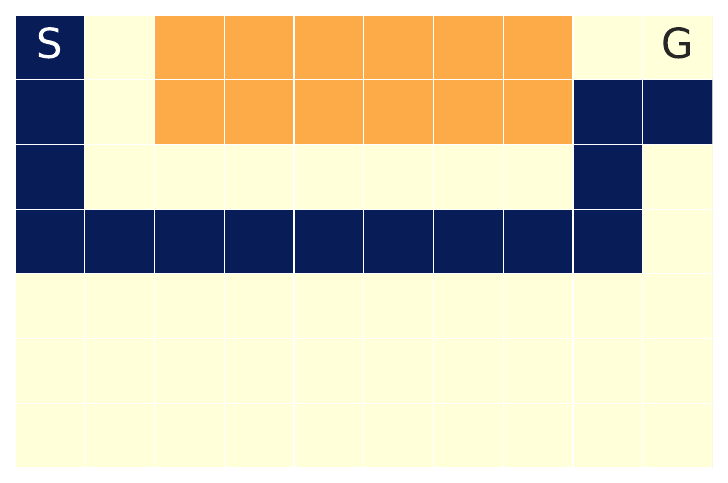}
    \includegraphics[width=.5\textwidth]{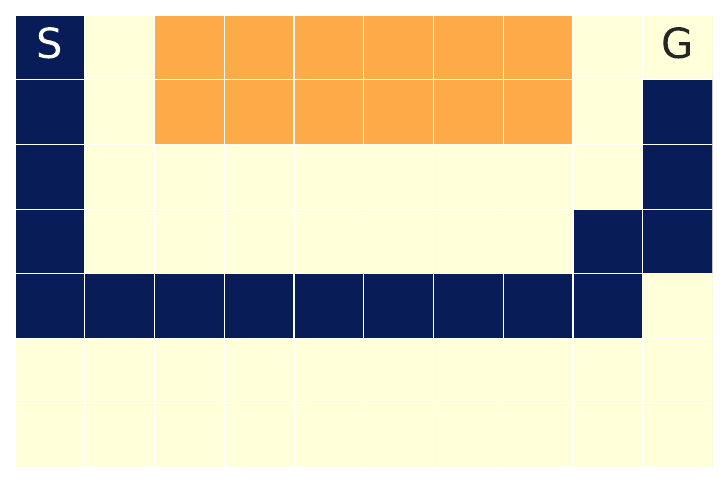}
    \includegraphics[width=.5\textwidth]{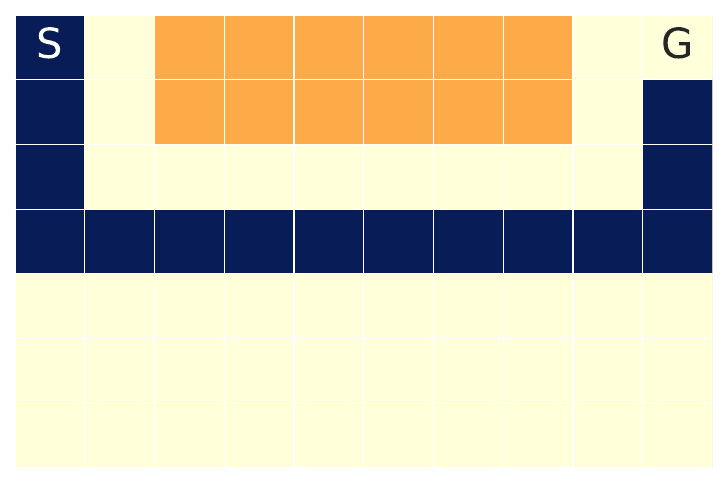}
    \caption{ACReL$_{0.04}$ ($\eta = 25$)}
\end{subfigure}%
\hspace{-2em}
\begin{subfigure}[c]{0.3\textwidth}\centering
    \includegraphics[width=.5\textwidth]{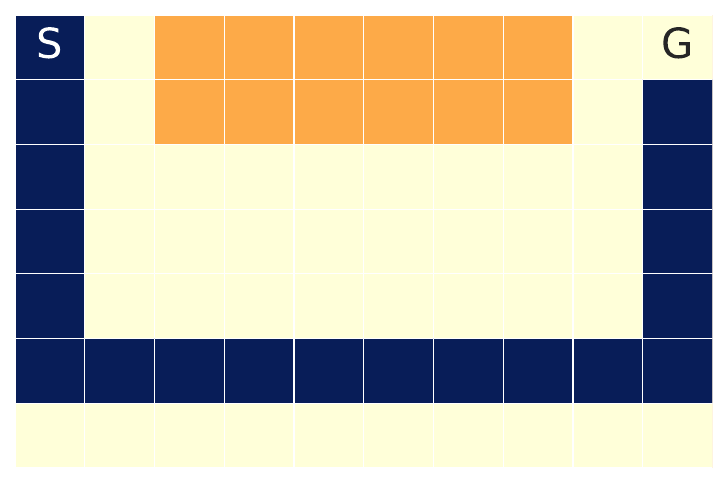}
    \includegraphics[width=.5\textwidth]{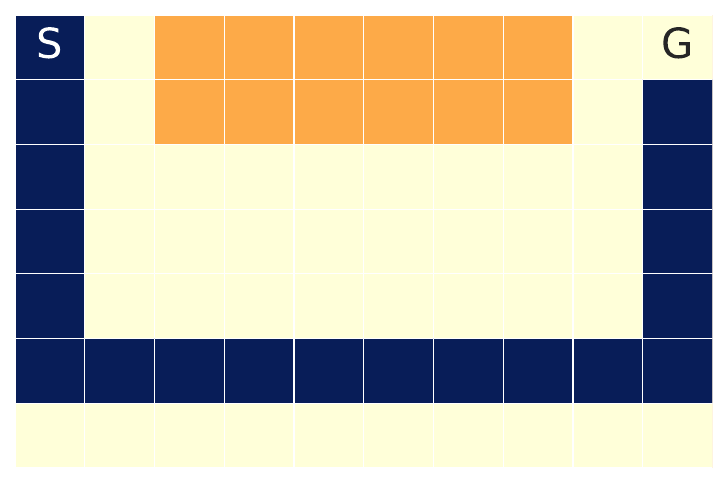}
    \includegraphics[width=.5\textwidth]{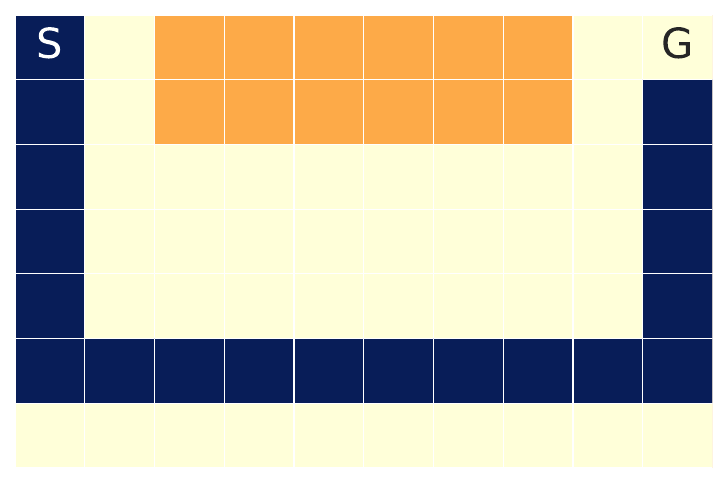}
    \includegraphics[width=.5\textwidth]{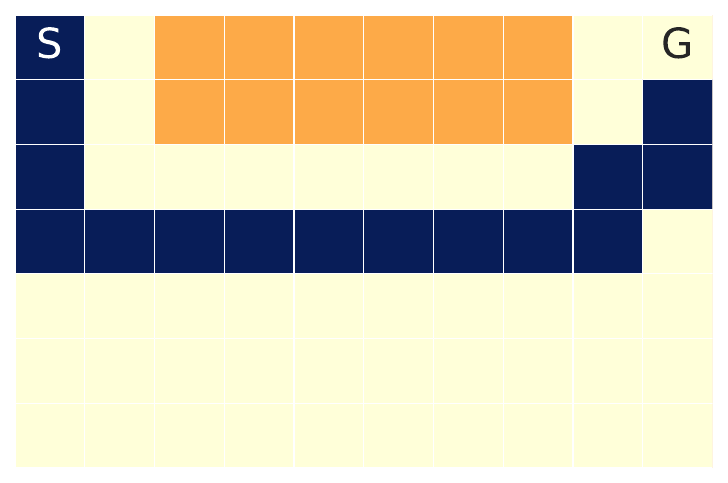}
    \includegraphics[width=.5\textwidth]{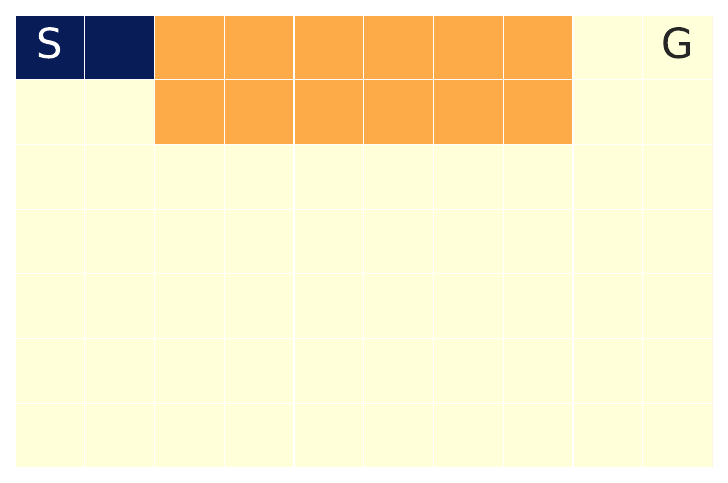}
    \includegraphics[width=.5\textwidth]{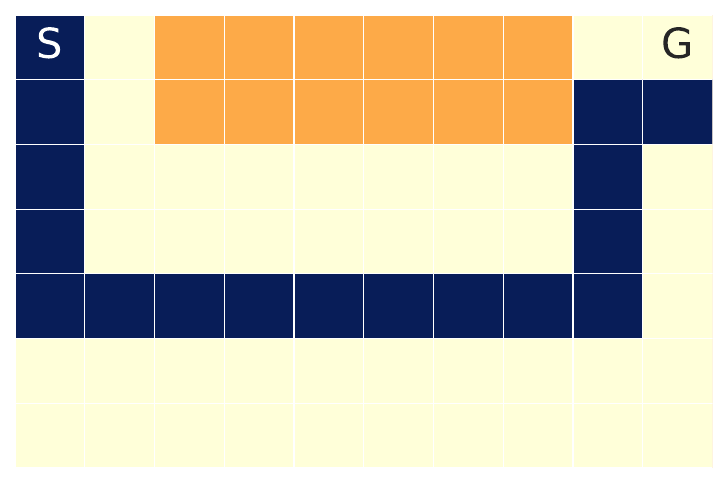}
    \includegraphics[width=.5\textwidth]{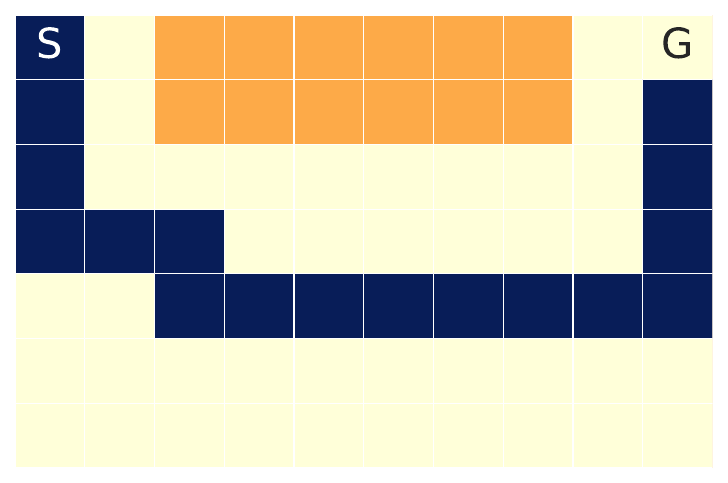}
    \includegraphics[width=.5\textwidth]{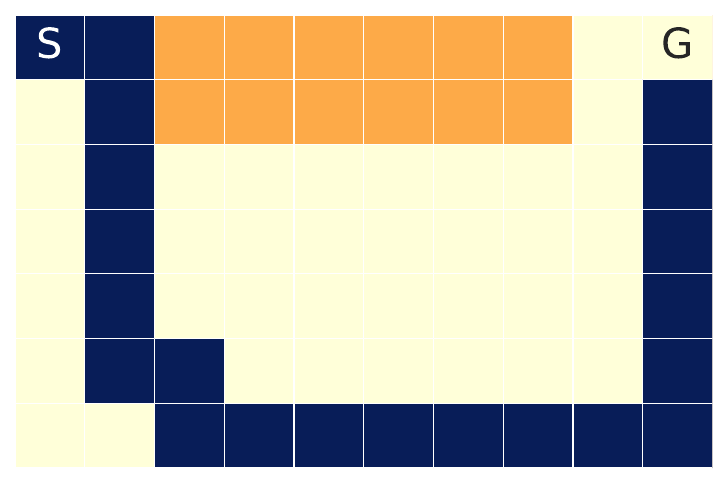}
    \includegraphics[width=.5\textwidth]{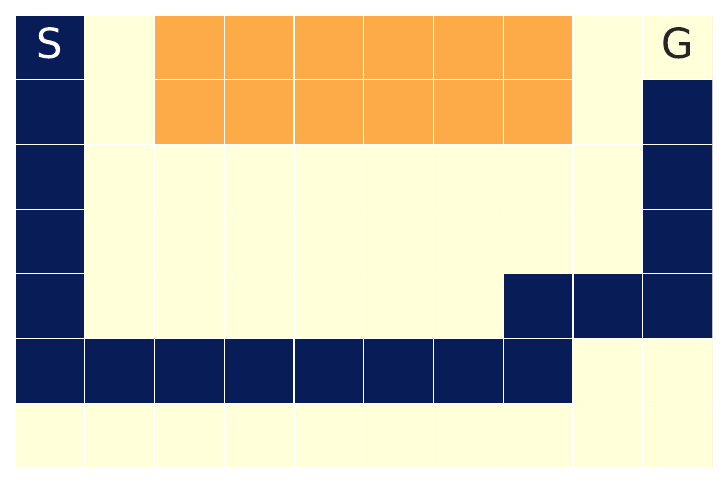}
    \includegraphics[width=.5\textwidth]{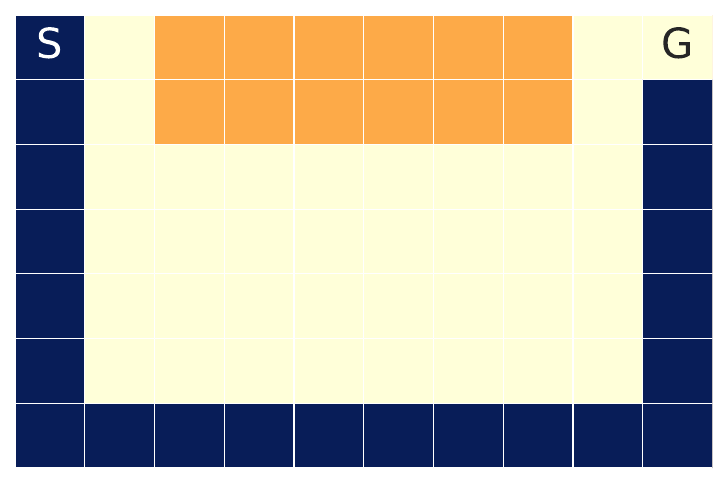}
    \caption{ACReL$_{0.01}$ ($\eta = 100$)}
\end{subfigure}
\caption{Exhaustive visualization of the ACReL policies collected to generate Figure~\ref{fig:averaged_seeds} (1st row). Each row figure displays ACReL policies learned with different budgets $\eta$ on the environment with $p=0.1$, for a random seed $S$.}
\label{fig:exhaustiveACREL}
\end{figure*}

\begin{figure*}
\centering
  \begin{subfigure}[c]{0.3\textwidth}\centering
    \myrowlabel{$S = 1$}
    \raisebox{-.5\height}{\includegraphics[width=.5\textwidth]{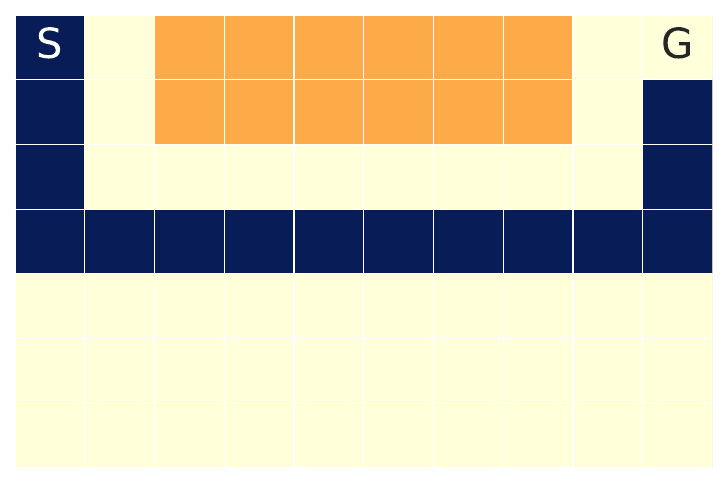}}\\
    \myrowlabel{$S = 2$}
    \raisebox{-.5\height}{\includegraphics[width=.5\textwidth]{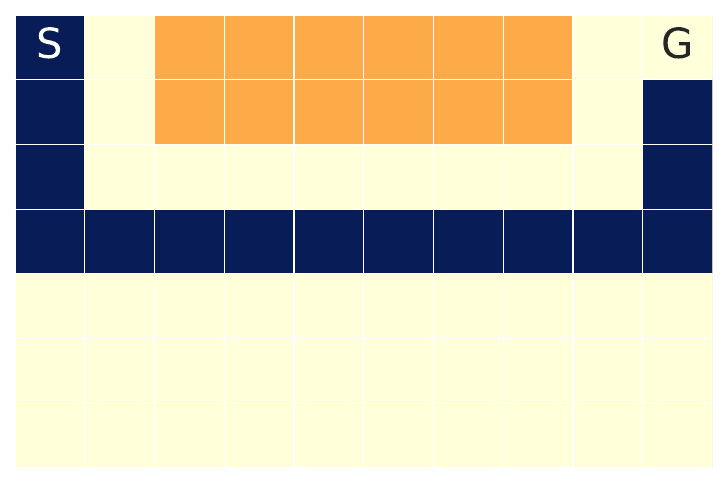}}\\
    \myrowlabel{$S = 3$}
    \raisebox{-.5\height}{\includegraphics[width=.5\textwidth]{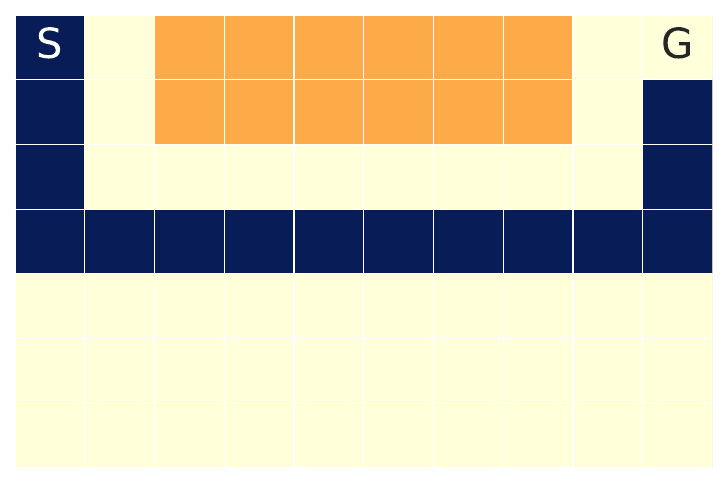}}\\
    \myrowlabel{$S = 4$}
    \raisebox{-.5\height}{\includegraphics[width=.5\textwidth]{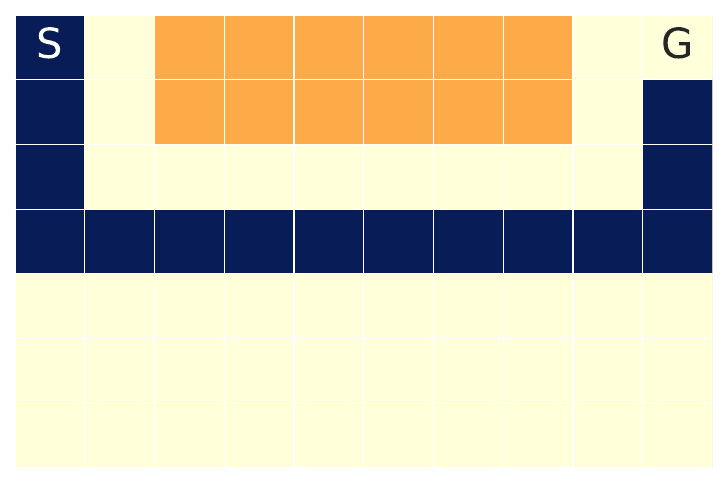}}\\
    \myrowlabel{$S = 5$}
    \raisebox{-.5\height}{\includegraphics[width=.5\textwidth]{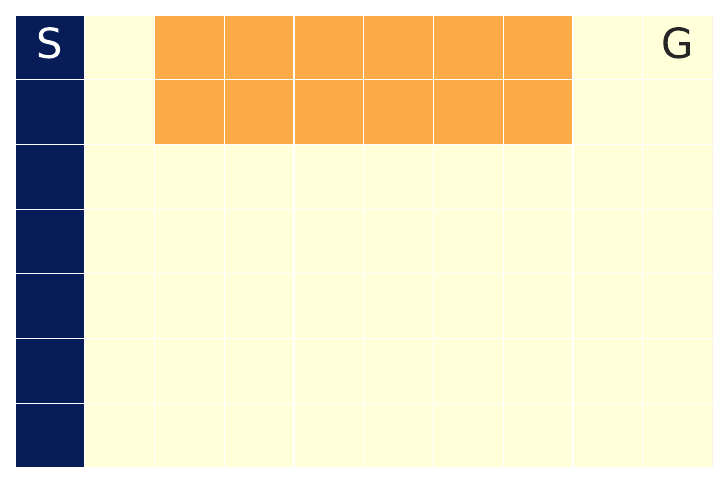}}\\
    \myrowlabel{$S = 6$}
    \raisebox{-.5\height}{\includegraphics[width=.5\textwidth]{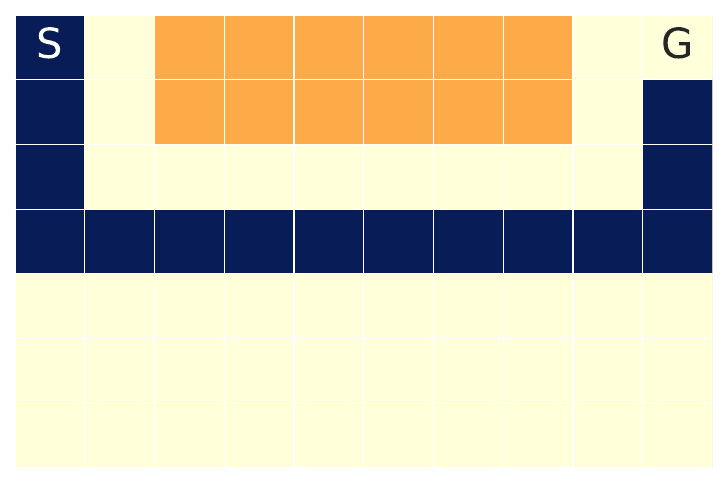}}\\
    \myrowlabel{$S = 7$}
    \raisebox{-.5\height}{\includegraphics[width=.5\textwidth]{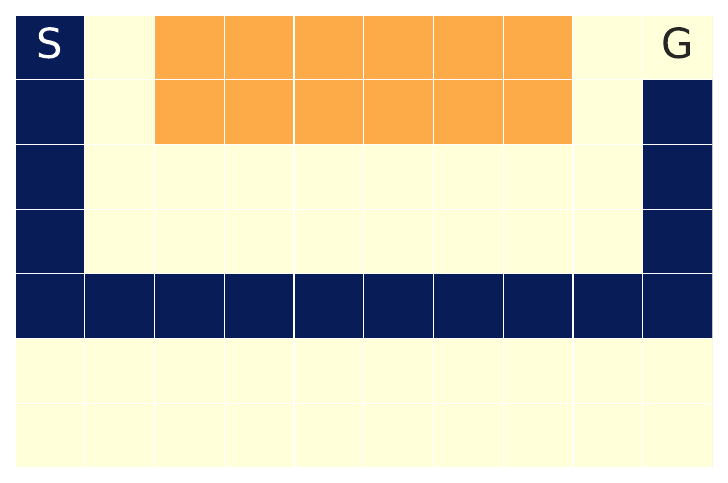}}\\
    \myrowlabel{$S = 8$}
    \raisebox{-.5\height}{\includegraphics[width=.5\textwidth]{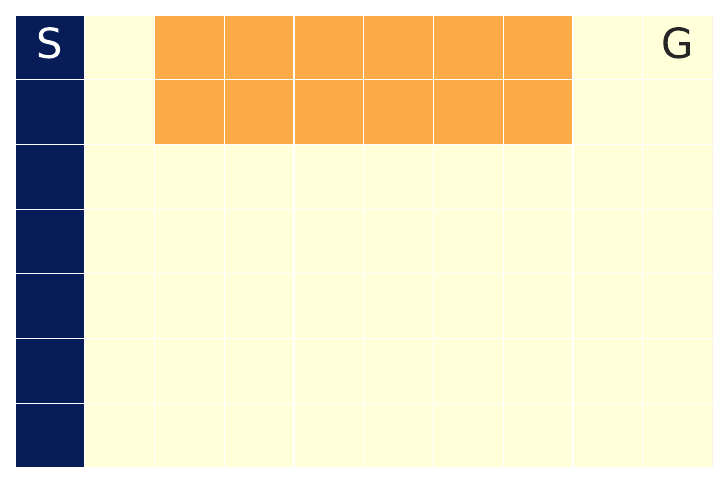}}\\
    \myrowlabel{$S = 9$}
    \raisebox{-.5\height}{\includegraphics[width=.5\textwidth]{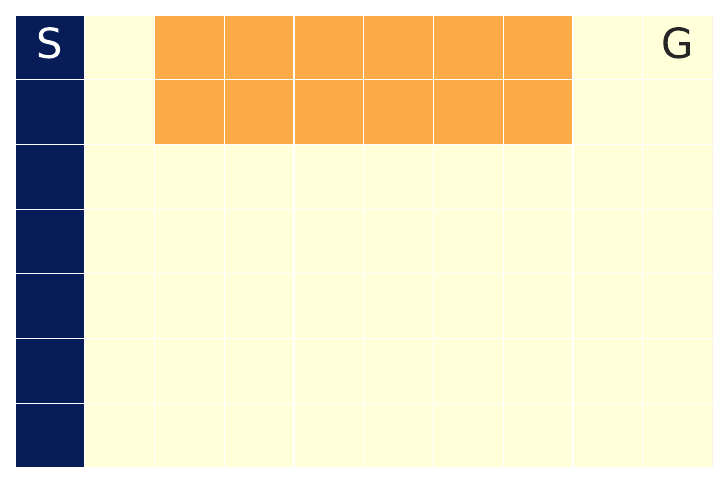}}\\
    \myrowlabel{$S = 10$}
    \raisebox{-.5\height}{\includegraphics[width=.5\textwidth]{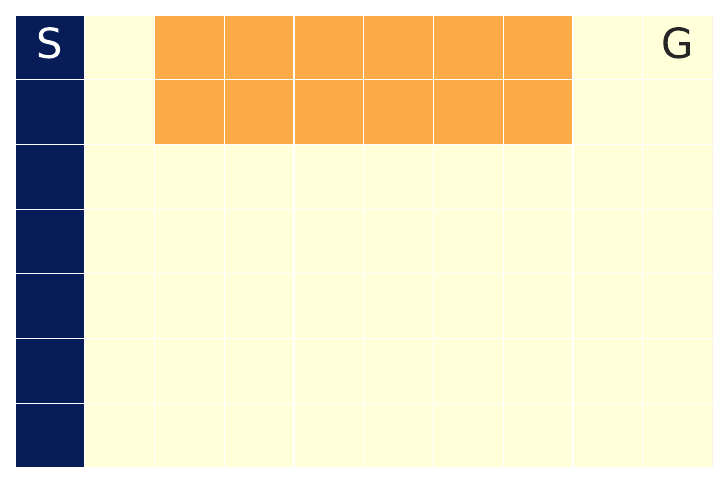}}\\
    \caption{IQN-$\text{CVaR}_{1}$}
\end{subfigure}%
\hspace{-2em}
\begin{subfigure}[c]{0.3\textwidth}\centering
    \includegraphics[width=.5\textwidth]{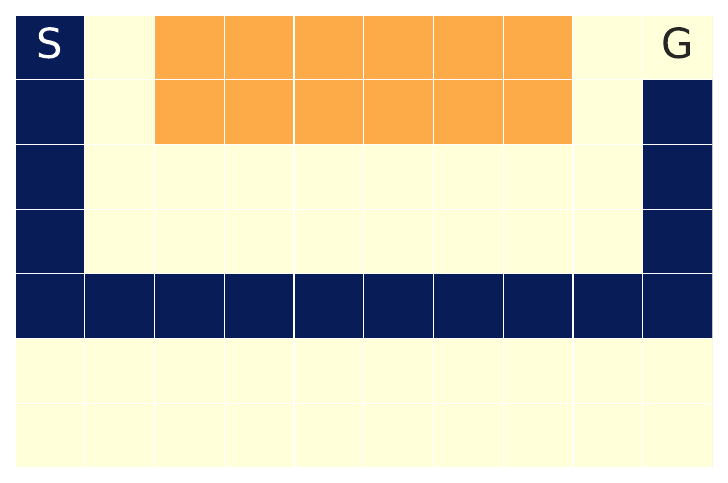}
    \includegraphics[width=.5\textwidth]{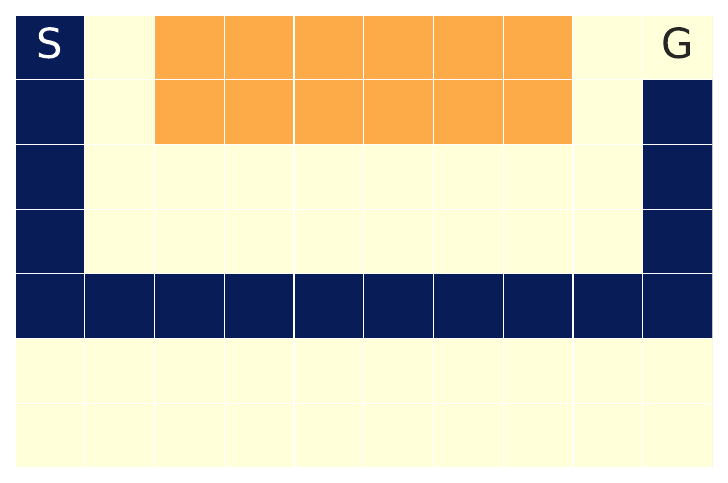}
    \includegraphics[width=.5\textwidth]{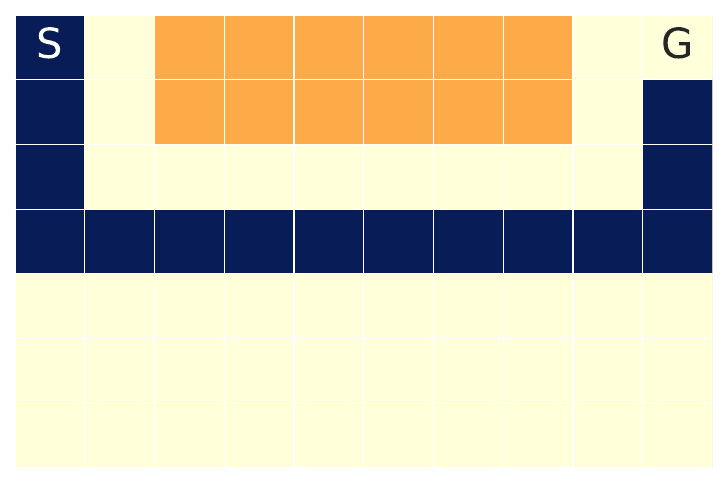}
    \includegraphics[width=.5\textwidth]{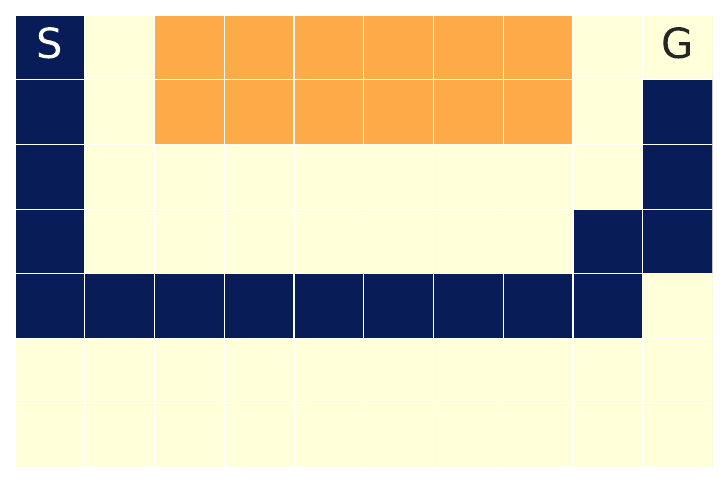}
    \includegraphics[width=.5\textwidth]{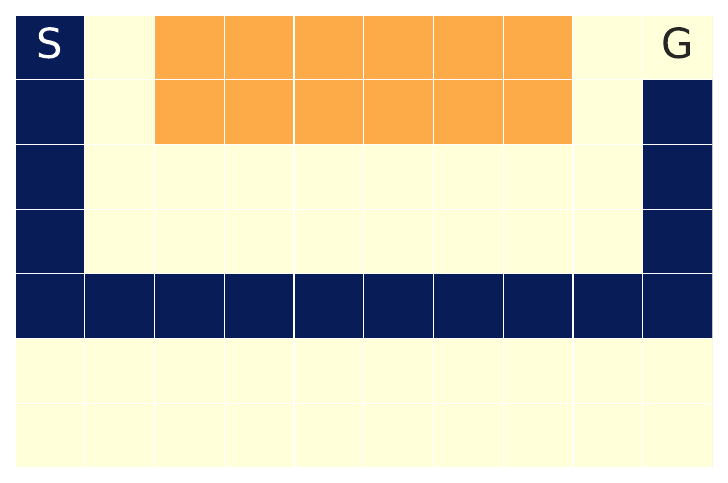}
    \includegraphics[width=.5\textwidth]{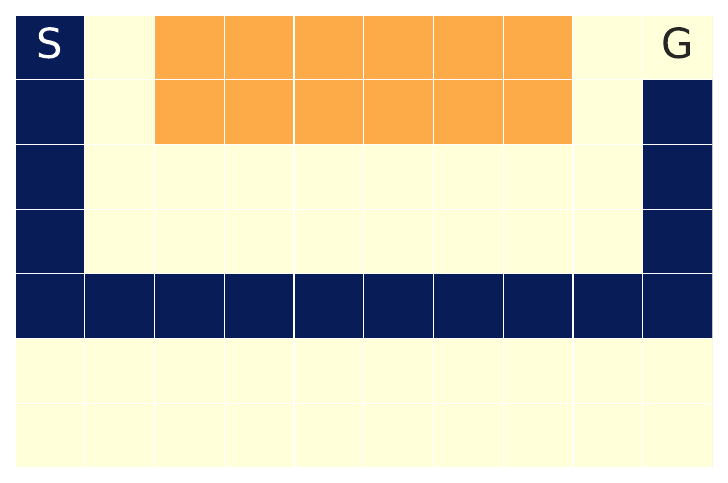}
    \includegraphics[width=.5\textwidth]{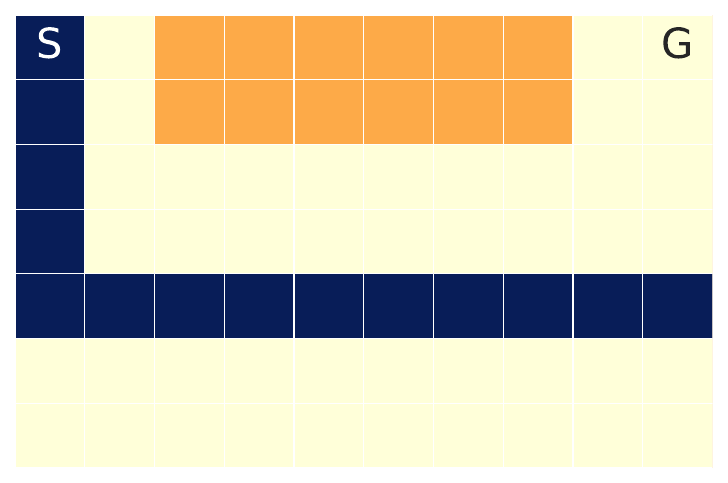}
    \includegraphics[width=.5\textwidth]{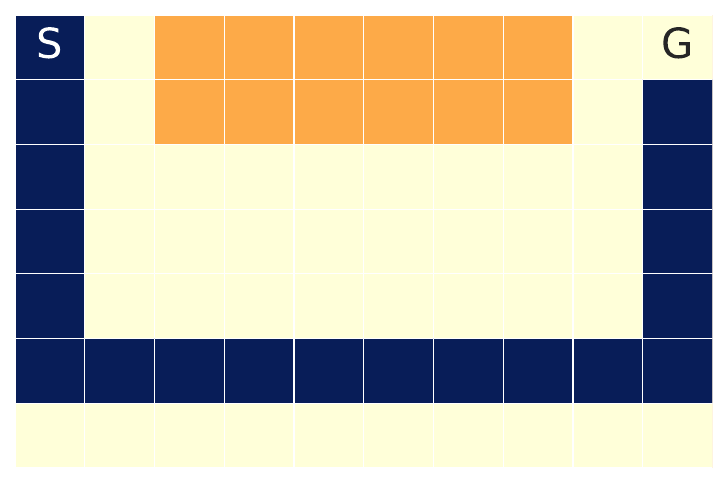}
    \includegraphics[width=.5\textwidth]{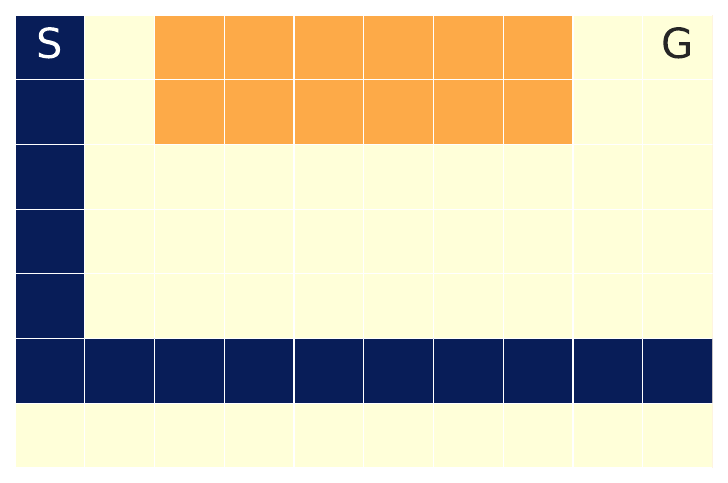}
    \includegraphics[width=.5\textwidth]{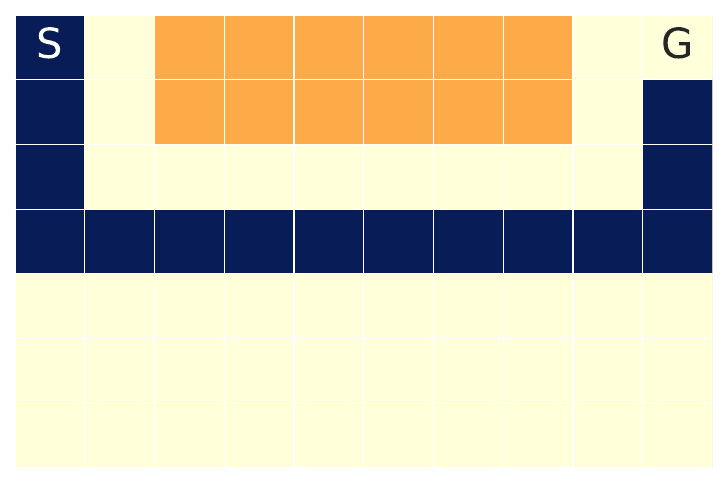}
    \caption{IQN-$\text{CVaR}_{0.04}$}
\end{subfigure}%
\hspace{-2em}
\begin{subfigure}[c]{0.3\textwidth}\centering
    \includegraphics[width=.5\textwidth]{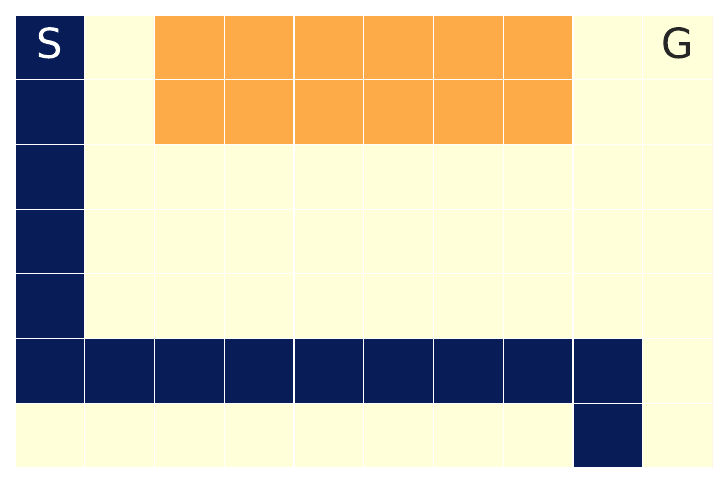}
    \includegraphics[width=.5\textwidth]{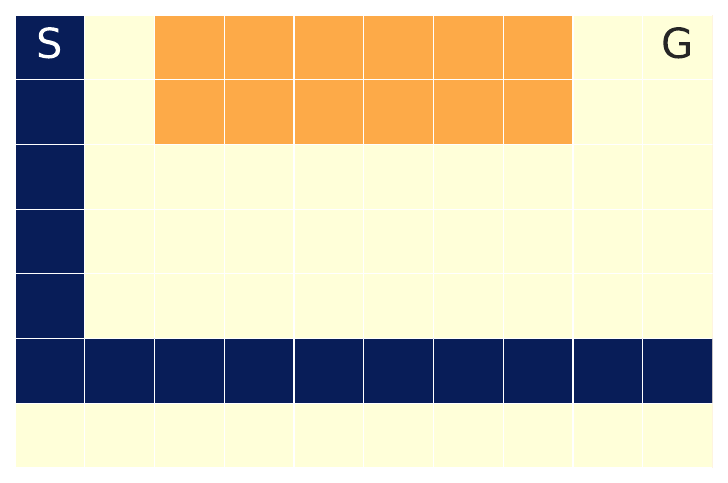}
    \includegraphics[width=.5\textwidth]{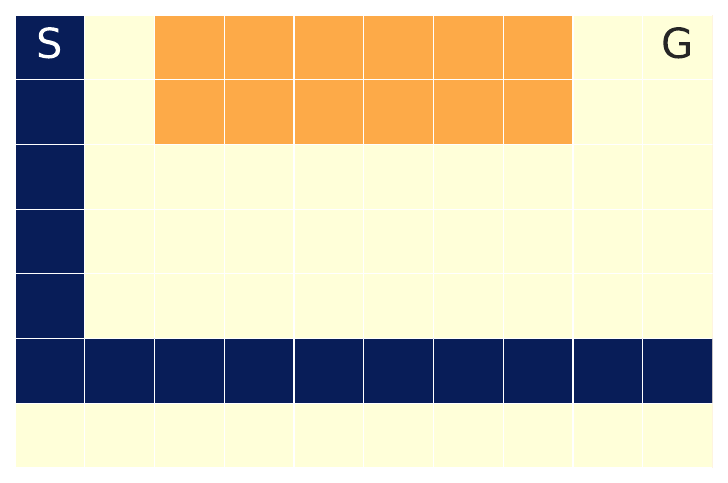}
    \includegraphics[width=.5\textwidth]{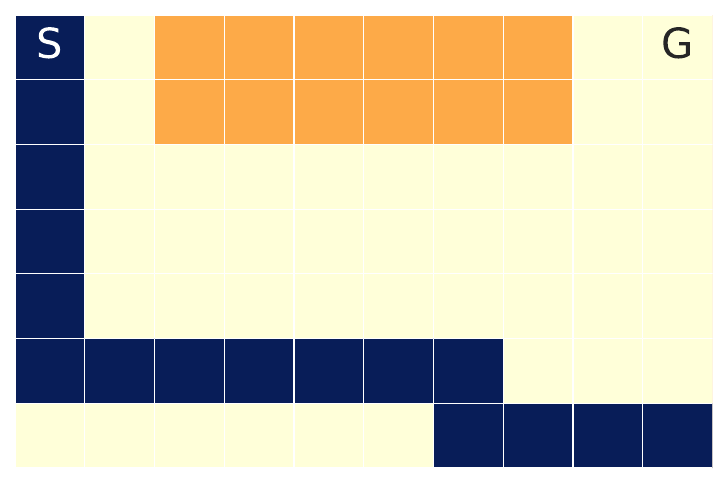}
    \includegraphics[width=.5\textwidth]{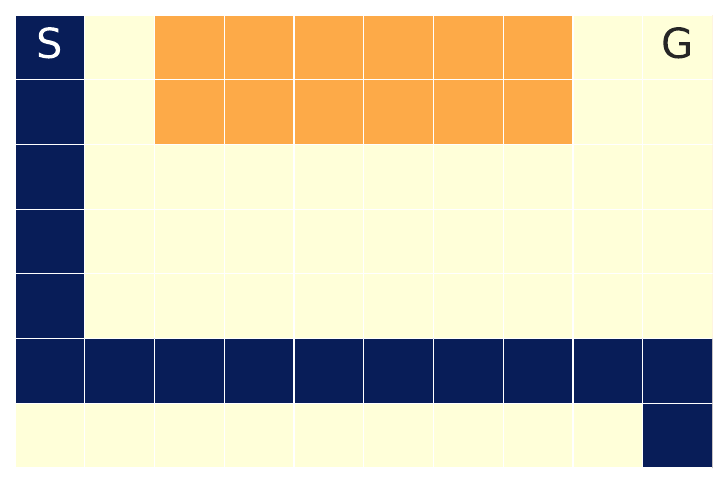}
    \includegraphics[width=.5\textwidth]{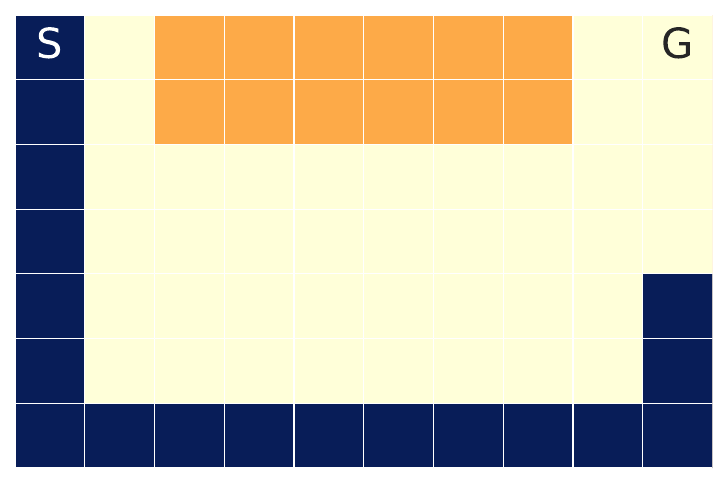}
    \includegraphics[width=.5\textwidth]{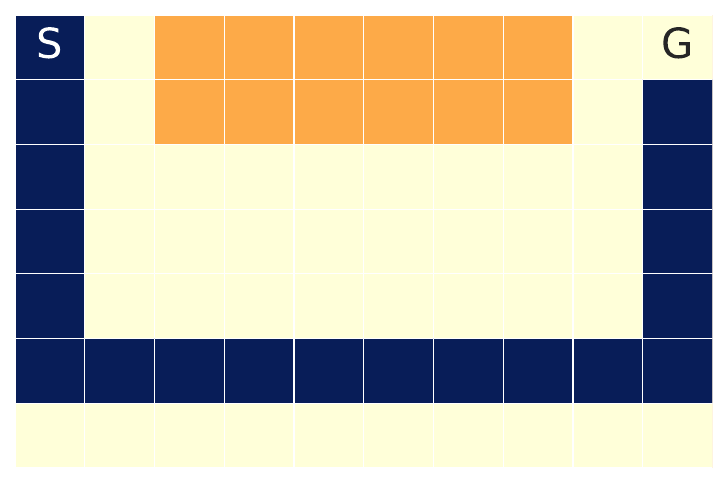}
    \includegraphics[width=.5\textwidth]{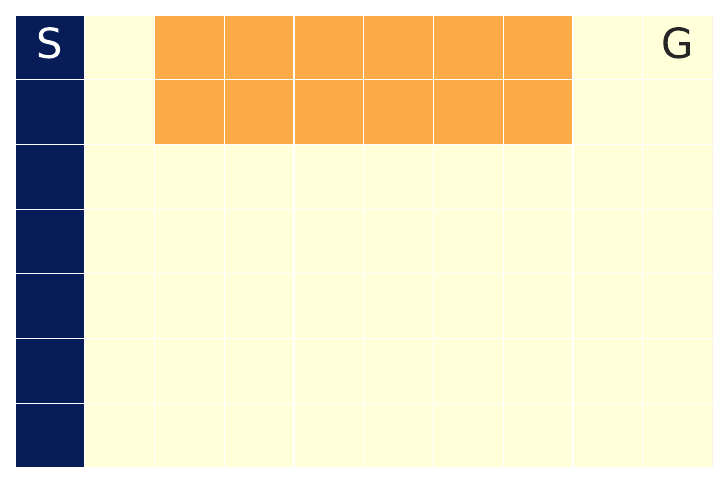}
    \includegraphics[width=.5\textwidth]{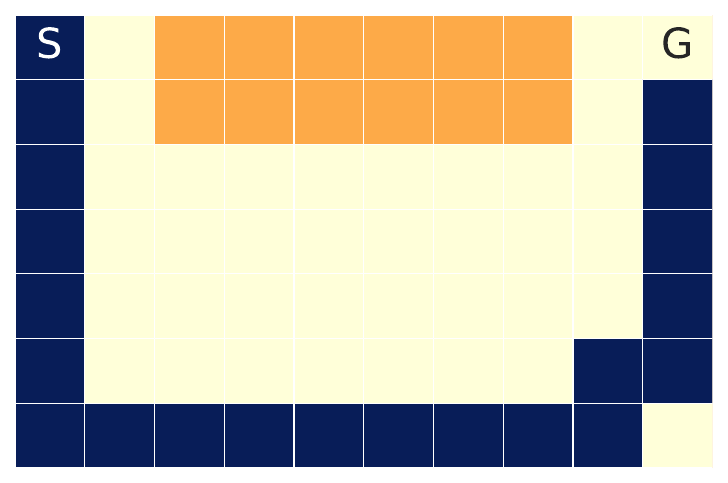}
    \includegraphics[width=.5\textwidth]{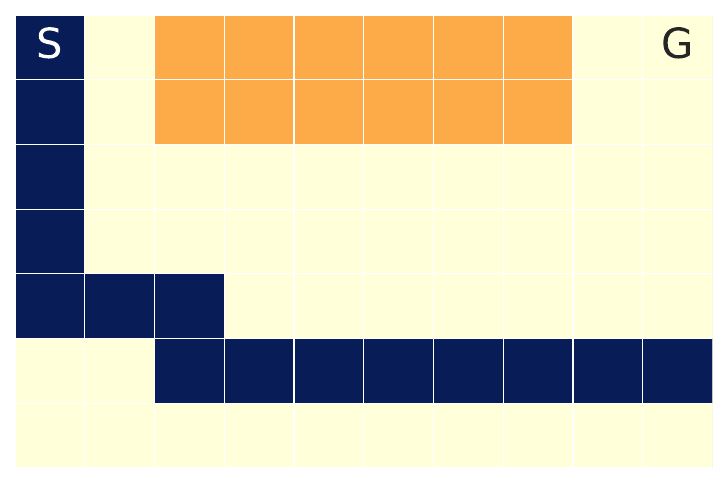}
    \caption{IQN-$\text{CVaR}_{0.01}$}
\end{subfigure}
\caption{Exhaustive visualization of the IQN-CVaR policies trained on $5M$ timesteps. 
These were averaged to generate Figure~\ref{fig:averaged_seeds} (2nd row). 
Each row figure displays policies learned with different levels $\text{CVaR}_{\alpha}$ on the environment with $p=0.1$, for a random seed $S$.}
\label{fig:exhaustiveIQN5M}
\end{figure*}

\begin{figure*}
\centering
  \begin{subfigure}[c]{0.3\textwidth}\centering
    \myrowlabel{$S = 1$}
    \raisebox{-.5\height}{\includegraphics[width=.5\textwidth]{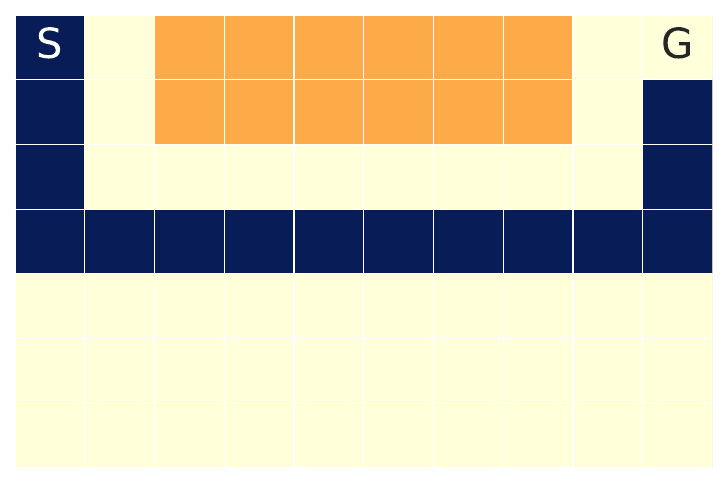}}\\
    \myrowlabel{$S = 2$}
    \raisebox{-.5\height}{\includegraphics[width=.5\textwidth]{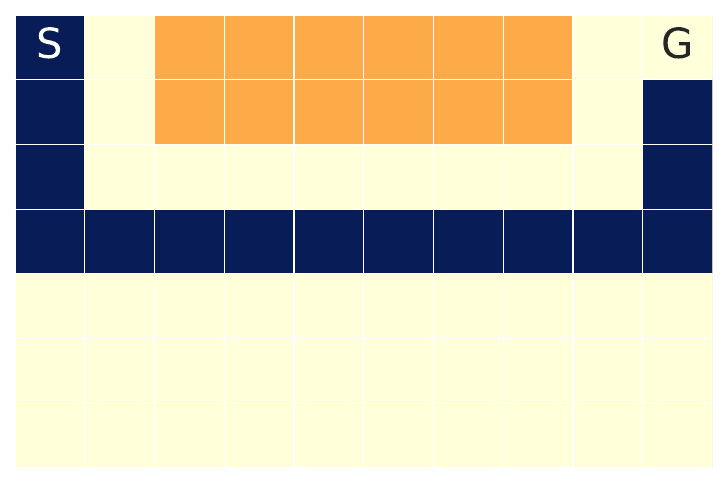}}\\
    \myrowlabel{$S = 3$}
    \raisebox{-.5\height}{\includegraphics[width=.5\textwidth]{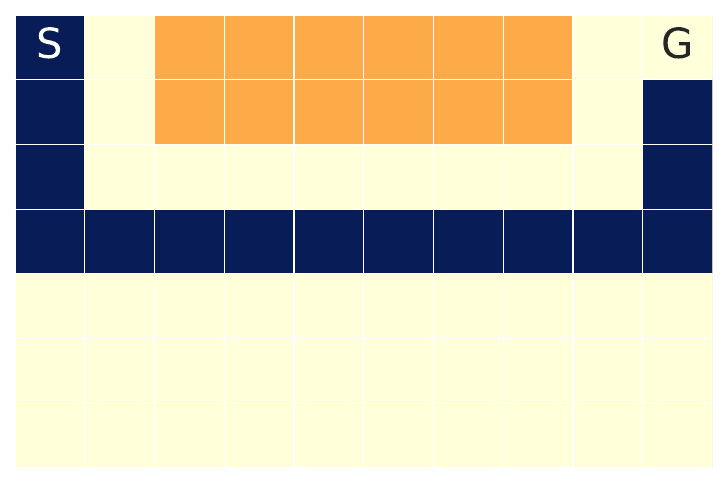}}\\
    \myrowlabel{$S = 4$}
    \raisebox{-.5\height}{\includegraphics[width=.5\textwidth]{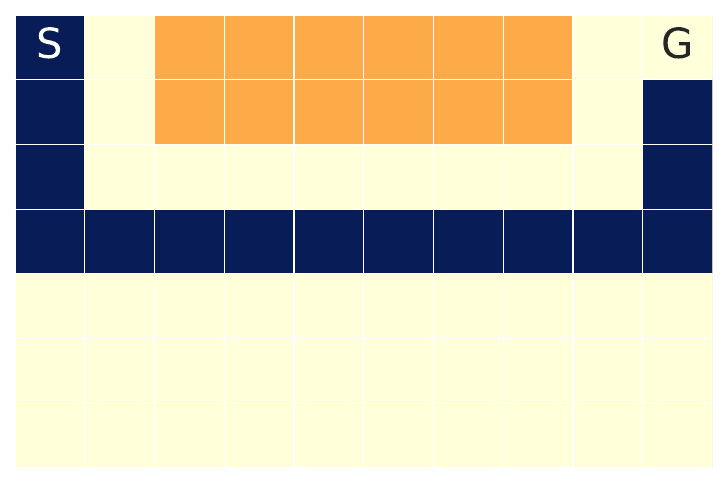}}\\
    \myrowlabel{$S = 5$}
    \raisebox{-.5\height}{\includegraphics[width=.5\textwidth]{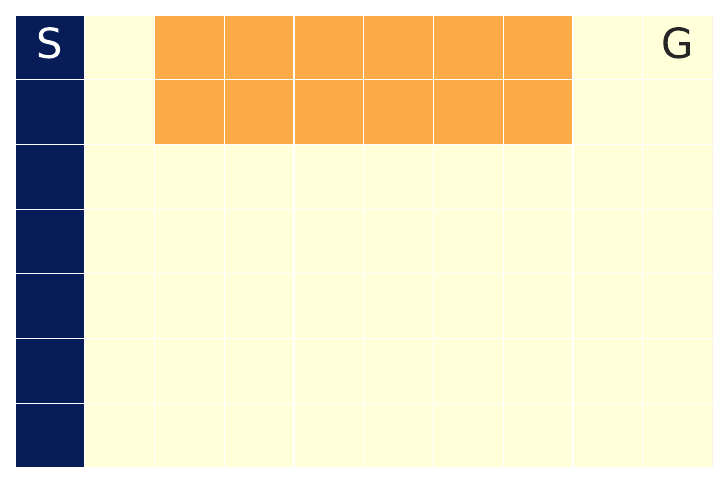}}\\
    \myrowlabel{$S = 6$}
    \raisebox{-.5\height}{\includegraphics[width=.5\textwidth]{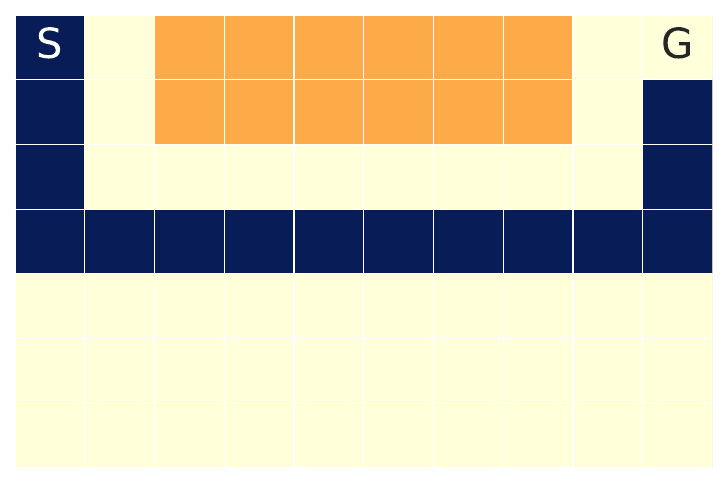}}\\
    \myrowlabel{$S = 7$}
    \raisebox{-.5\height}{\includegraphics[width=.5\textwidth]{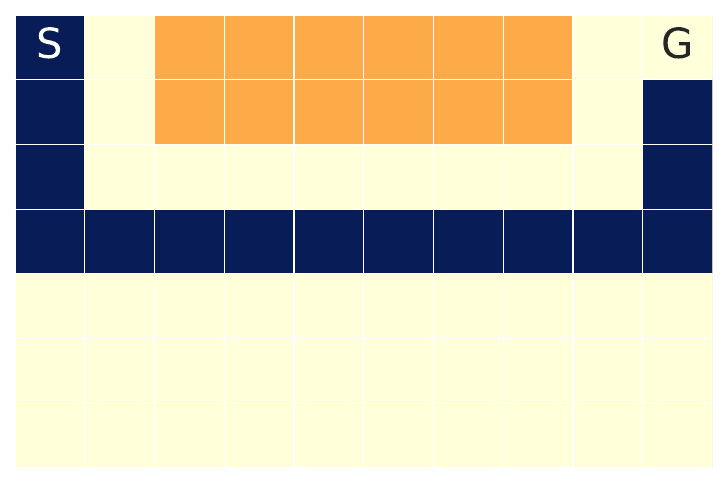}}\\
    \myrowlabel{$S = 8$}
    \raisebox{-.5\height}{\includegraphics[width=.5\textwidth]{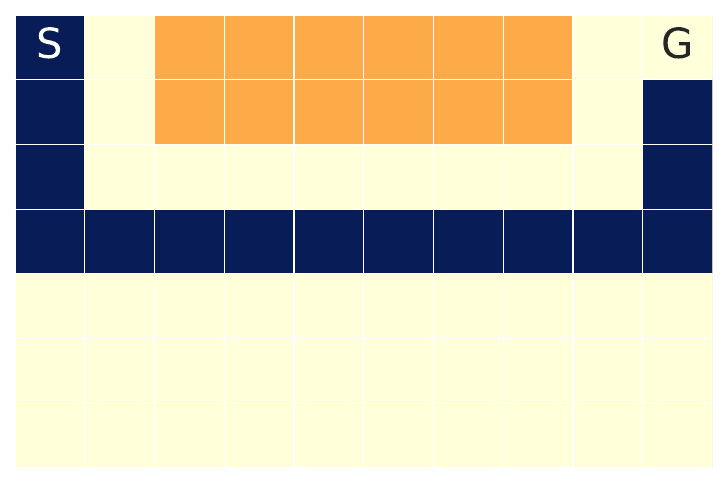}}\\
    \myrowlabel{$S = 9$}
    \raisebox{-.5\height}{\includegraphics[width=.5\textwidth]{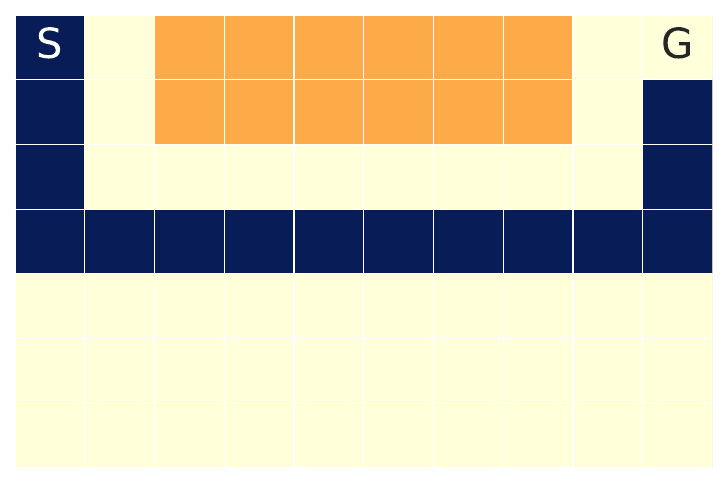}}\\
    \myrowlabel{$S = 10$}
    \raisebox{-.5\height}{\includegraphics[width=.5\textwidth]{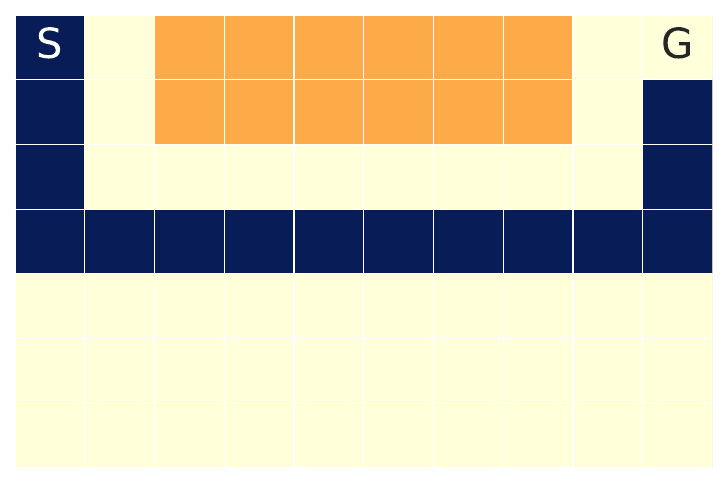}}\\
    \caption{IQN-$\text{CVaR}_{1}$}
\end{subfigure}%
\hspace{-2em}
\begin{subfigure}[c]{0.3\textwidth}\centering
    \includegraphics[width=.5\textwidth]{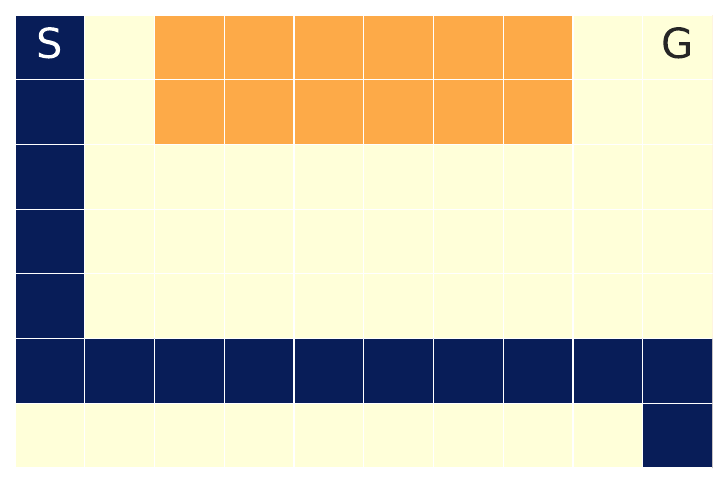}
    \includegraphics[width=.5\textwidth]{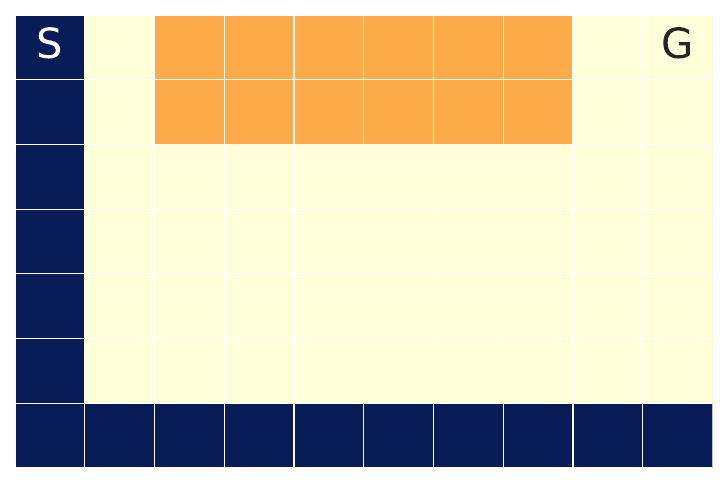}
    \includegraphics[width=.5\textwidth]{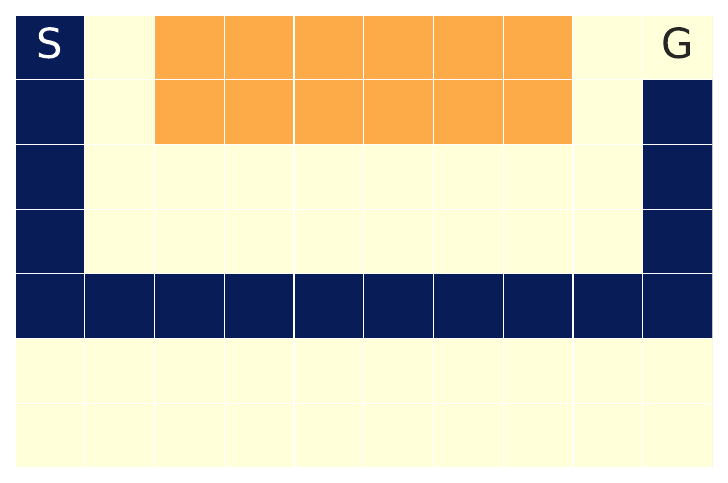}
    \includegraphics[width=.5\textwidth]{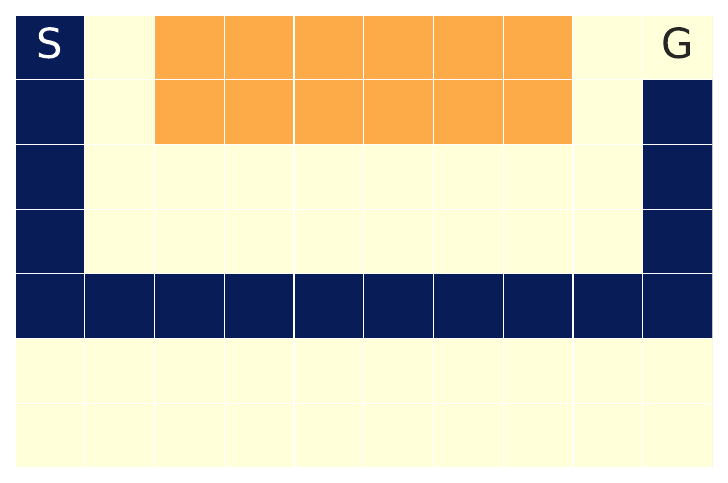}
    \includegraphics[width=.5\textwidth]{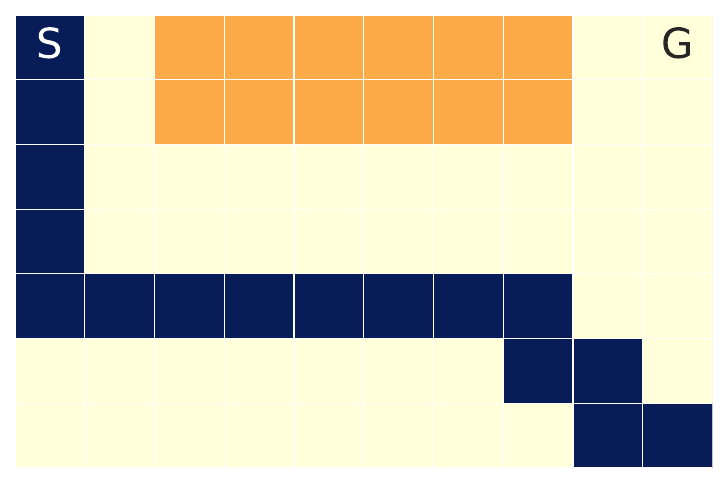}
    \includegraphics[width=.5\textwidth]{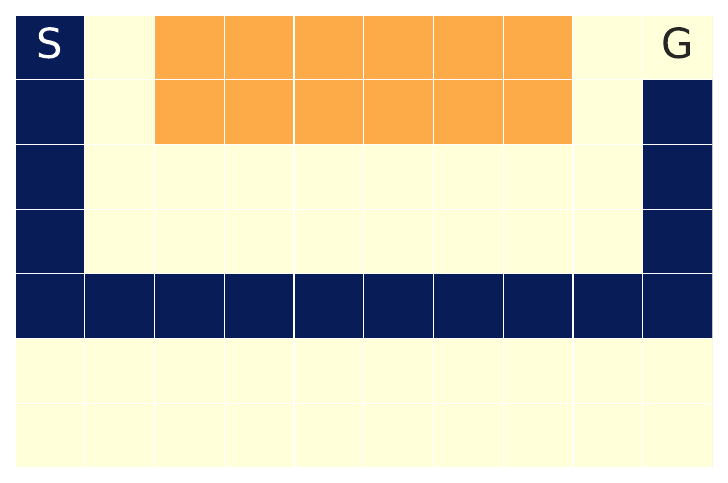}
    \includegraphics[width=.5\textwidth]{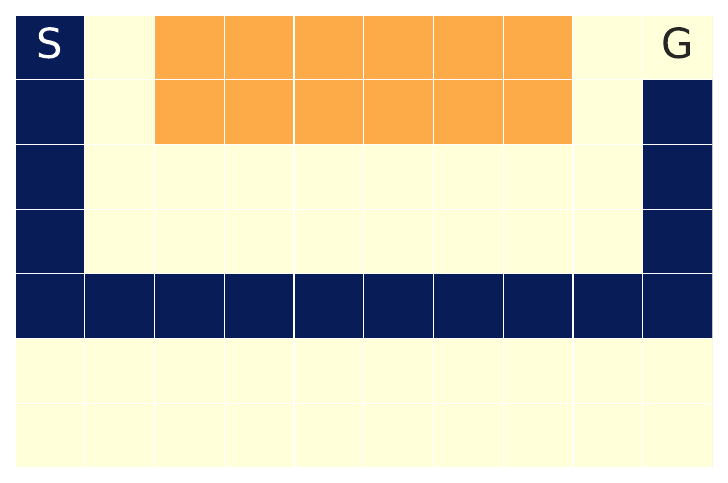}
    \includegraphics[width=.5\textwidth]{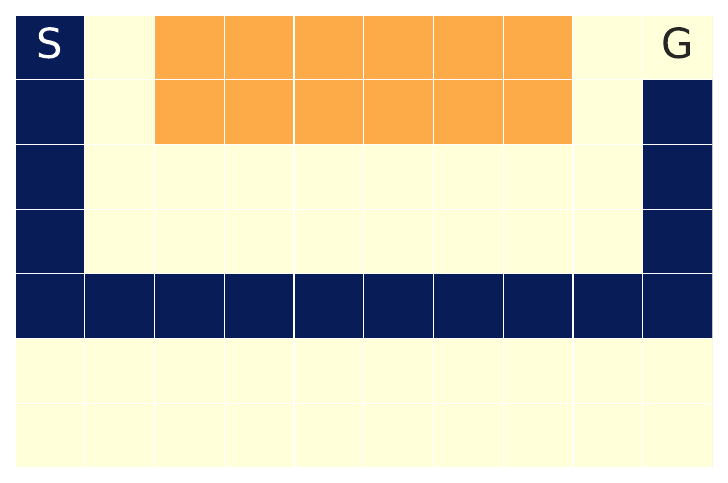}
    \includegraphics[width=.5\textwidth]{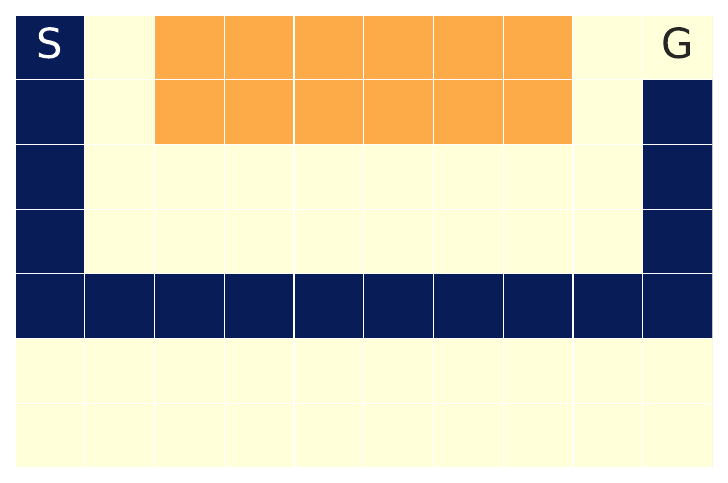}
    \includegraphics[width=.5\textwidth]{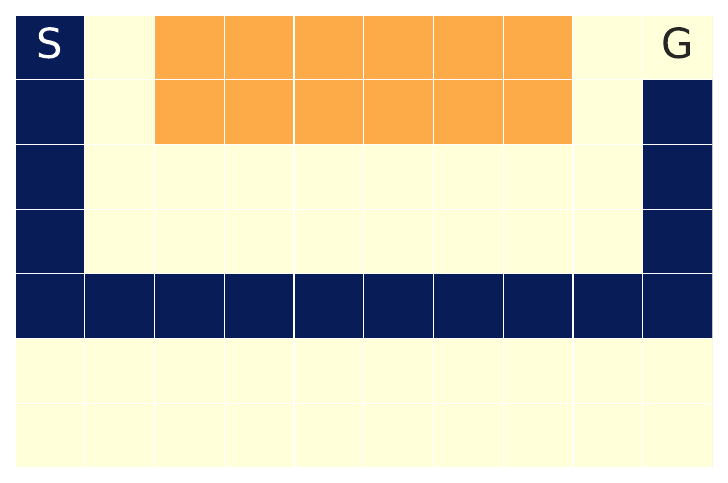}
    \caption{IQN-$\text{CVaR}_{0.04}$}
\end{subfigure}%
\hspace{-2em}
\begin{subfigure}[c]{0.3\textwidth}\centering
    \includegraphics[width=.5\textwidth]{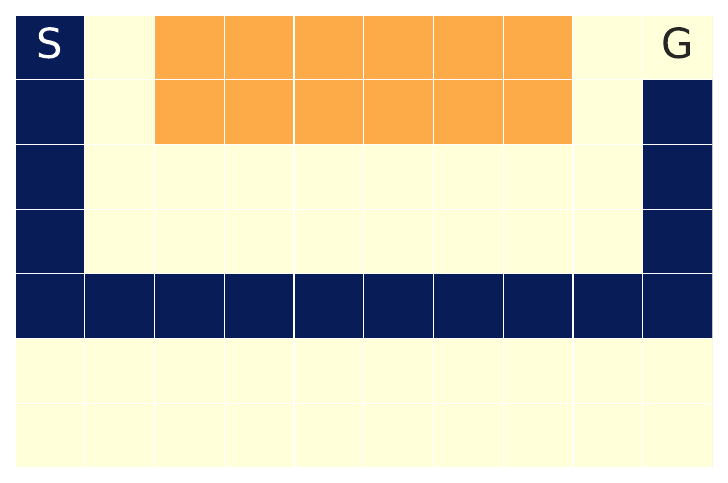}
    \includegraphics[width=.5\textwidth]{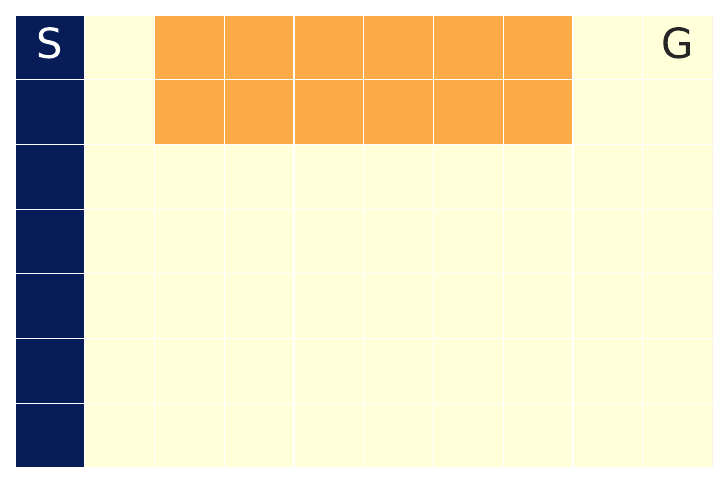}
    \includegraphics[width=.5\textwidth]{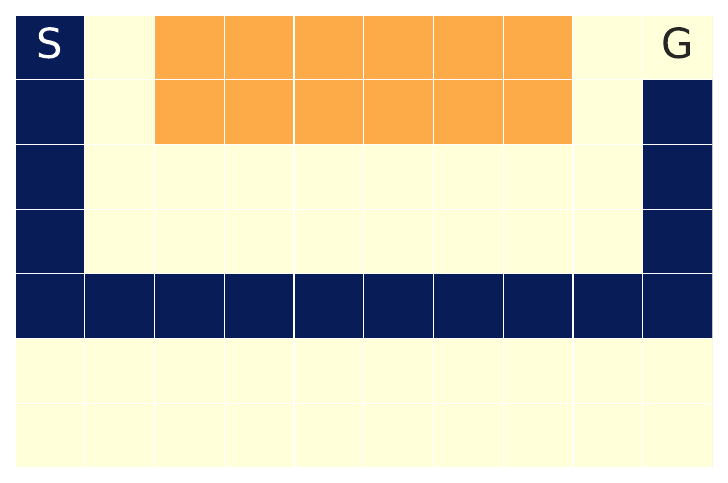}
    \includegraphics[width=.5\textwidth]{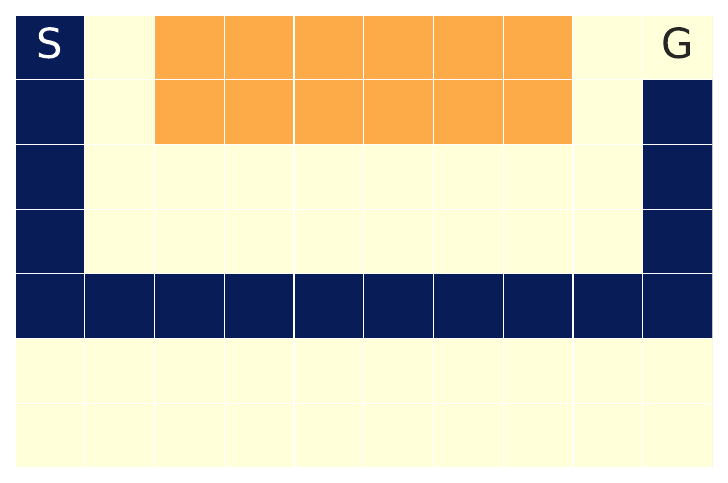}
    \includegraphics[width=.5\textwidth]{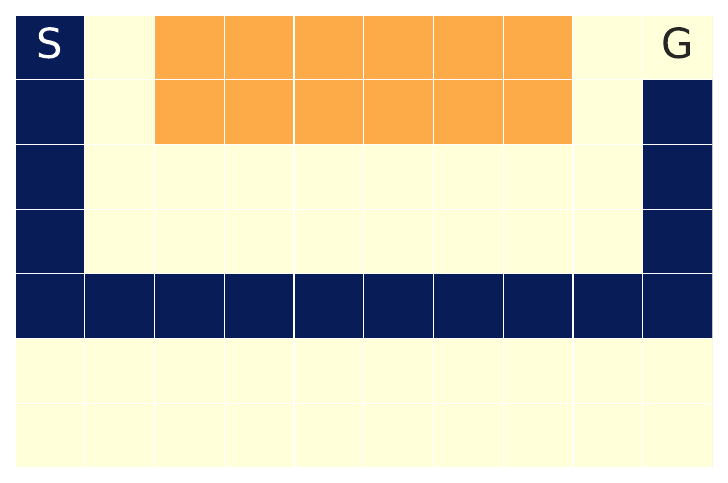}
    \includegraphics[width=.5\textwidth]{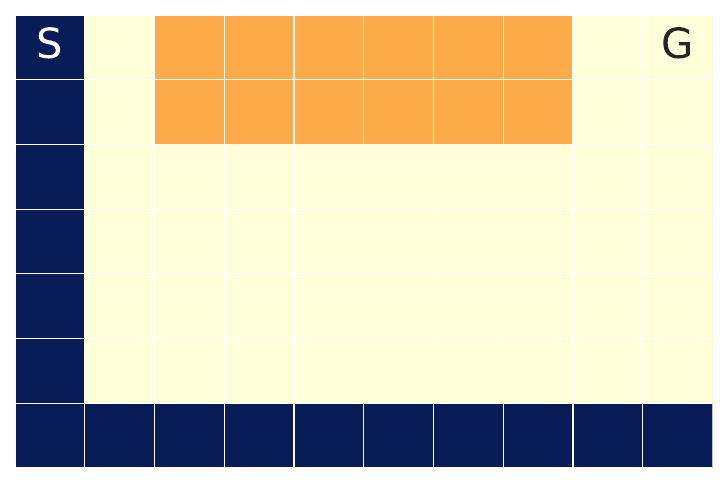}
    \includegraphics[width=.5\textwidth]{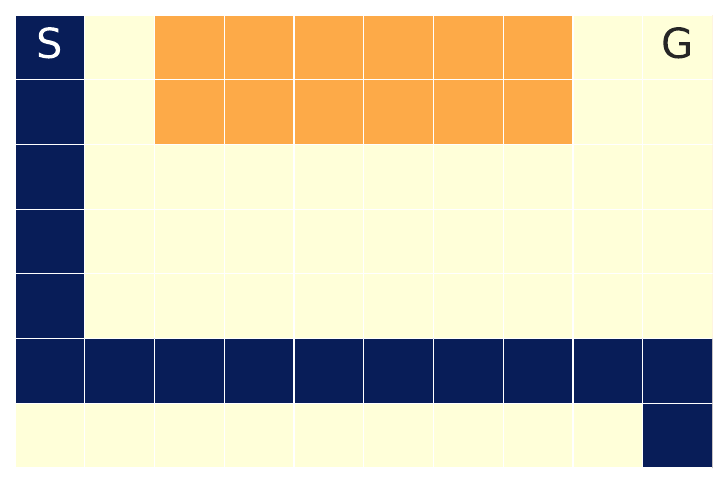}
    \includegraphics[width=.5\textwidth]{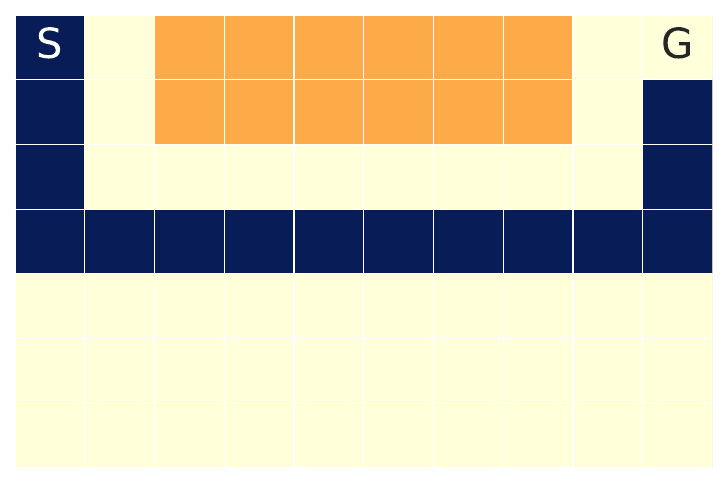}
    \includegraphics[width=.5\textwidth]{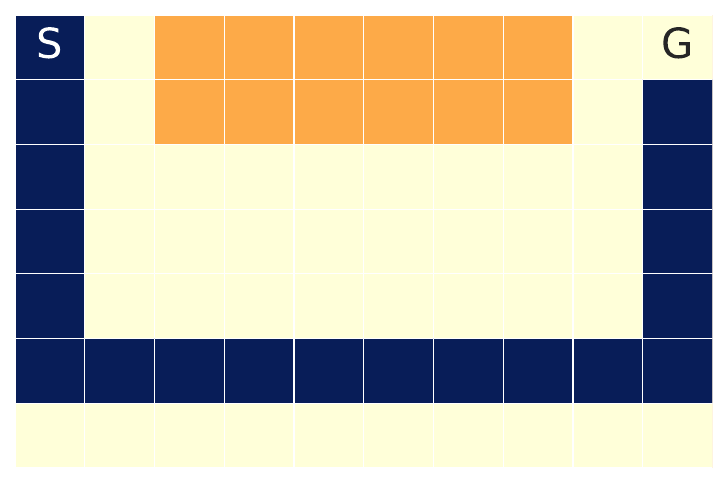}
    \includegraphics[width=.5\textwidth]{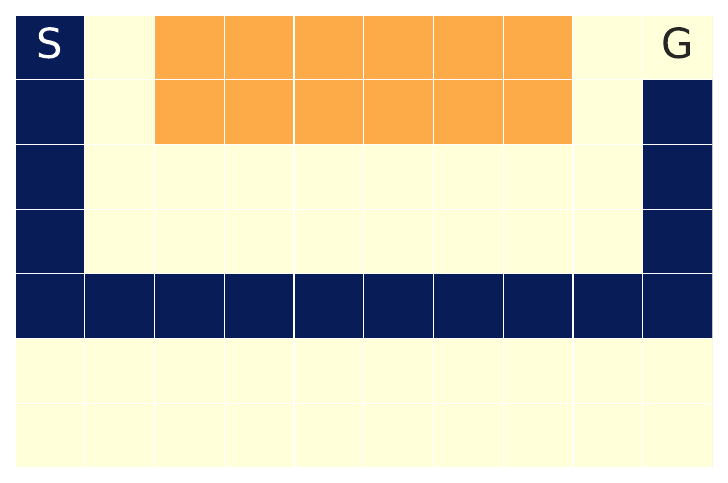}
    \caption{IQN-$\text{CVaR}_{0.01}$}
\end{subfigure}
\caption{Exhaustive visualization of the IQN-CVaR policies trained on $10M$ timesteps. 
Each row figure displays policies learned with different levels $\text{CVaR}_{\alpha}$ on the environment with $p=0.1$, for a random seed $S$.}
\label{fig:exhaustiveIQN10M}
\end{figure*}

While we reported learned policies averaged over $n=10$ random seeds in the main paper, we now present each separate training run's results.
The agent policies obtained with ACReL are displayed in Figure~\ref{fig:exhaustiveACREL} while those for IQN-CVaR are in Figure~\ref{fig:exhaustiveIQN5M}.
In Figure~\ref{fig:exhaustiveIQN10M}, we also report IQN-CVaR's results when trained for $10M$ steps, to view whether or not IQN-CVaR's convergence instabilities may be solved by simply increasing the training duration.

Three different training results can be observed on the individual training run's learned policies.
A first pattern, is the U-shaped trajectories that is obtained when the policy successfully reached the goal state by going around the lava obstacle.
This first and most common pattern contains the trajectories which observe the most rewards, as any trajectory that reaches the goal state is guaranteed to obtain a sum of total rewards $R = \sum_t^T r_t \geq -0.4$, corresponding to the cost of 40 steps (-1.4) and the goal reward of +1.
A second pattern is the one where the agent simply wastes its time reaching a fixed wall tile and then staying in it for the remainder of the episode.
This pattern is guaranteed to produce a sum of rewards $R = -1.4$, corresponding to paying the step cost $-0.035$ for the maximum number of allowed steps $40$.
A third noticeable pattern is the one where the agent learns to immediately throw themselves in the lava obstacle.
This pattern is guaranteed to obtain a sum of rewards $R = -1.07$, corresponding to the cost of the two steps required to reach the lava and the lava reward of -1.

Intuitively, depending on whether or not the agent's exploration reached the goal state a sufficient number of times during training, one should expect learned policies to either harbor the first or third pattern, as the second one is the least rewarding.
We accordingly refer to the second pattern as a convergence failure.
Table \ref{tab:failures_to_converge} displays the number of convergence failures over the 10 considered seeds. 

\begin{table}
\centering
\resizebox{0.5 \textwidth}{!}{%
\begin{tabular}{l|c|c|c|}
\cline{2-4}
                                           & \multicolumn{1}{l|}{$\text{CVaR}_{1}$} & \multicolumn{1}{l|}{$\text{CVaR}_{0.04}$} & \multicolumn{1}{l|}{$\text{CVaR}_{0.01}$} \\ \hline
\multicolumn{1}{|l|}{ACReL}                & 2                               & 1                                  & 1                                  \\ \hline
\multicolumn{1}{|l|}{IQN-CVaR (5M steps)}  & 4                               & 2                                  & 8                                  \\ \hline
\multicolumn{1}{|l|}{IQN-CVaR (10M steps)} & 1                               & 3                                  & 3                                  \\ \hline
\end{tabular}%
}
\caption{Convergence failure count over 10 different seeds for ACReL and IQN-CVaR (trained over $5M$ and $10M$ steps) (computed from Figures~\ref{fig:exhaustiveACREL}, \ref{fig:exhaustiveIQN5M}, and \ref{fig:exhaustiveIQN10M})}
\label{tab:failures_to_converge}
\end{table}

We first note that ACReL is less prone to convergence failure than IQN-CVaR under a similar training regime of $5M$ steps.
Although increasing the number of steps to $10M$ for IQN-CVaR does seem to alleviate the convergence failure rate, it still does not manage to outperform ACReL with twice the training steps.
We hypothesize that this may be a symptom of IQN-CVaR's lack of theoretical convergence guarantee for risk-sensitive objectives or simply a result of the environment's intrinsic exploration difficulty.
Further experiments would be needed to explain this performance difference and we leave them as future work.


\subsection{Harder environment setting}
\label{harder_task}
In this section, we provide additional results on the evaluation of our method on a different configuration of the environment. 
In the main paper, the environment is stochastic with probability ($p = 0.1$) for the agent to perform a random action instead of the chosen action $a_t$. 
Here, we consider a different configuration where the probability of random action is set to $p=0.05$. 
The considered budgets for the adversary remain the same ($\eta \in \{1, 25, 100\}$ respectively corresponding to learning a CVaR$_\alpha$ of  $\alpha \in \{1, 0.04, 0.01\}$). 
This task is harder for the adversary because it has the same budget but must perform bigger perturbations to influence the agent's trajectories since the stochasticity of the environment is two times lower.

\begin{figure*}
\centering
\begin{subfigure}[t]{0.3 \linewidth }
    \includegraphics[width=\textwidth]{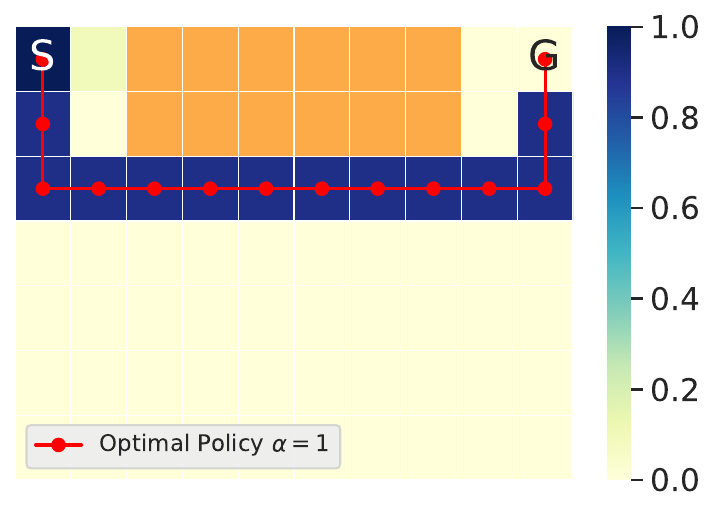}
    \caption{ACReL$_1$ $(\eta = 1)$}
    \label{fig:averaged_seeds_0005}
    \includegraphics[width=\textwidth]{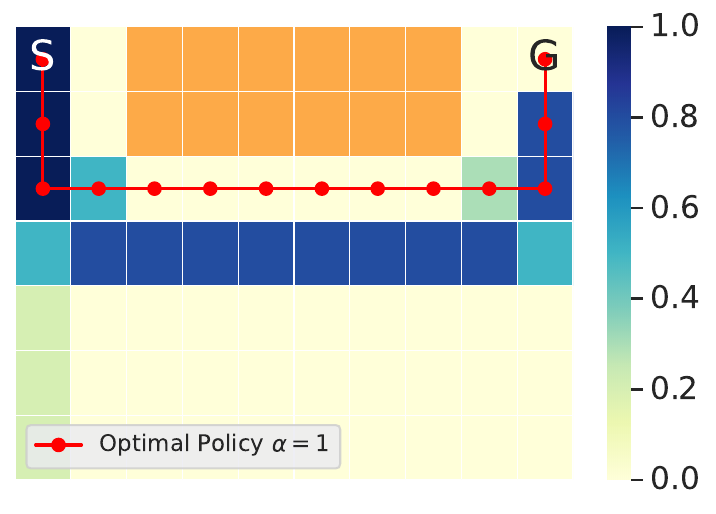}
    \caption{IQN-CVaR$_{1}$}
\end{subfigure}\hfill
\begin{subfigure}[t]{0.3 \linewidth}
    \includegraphics[width=\textwidth]{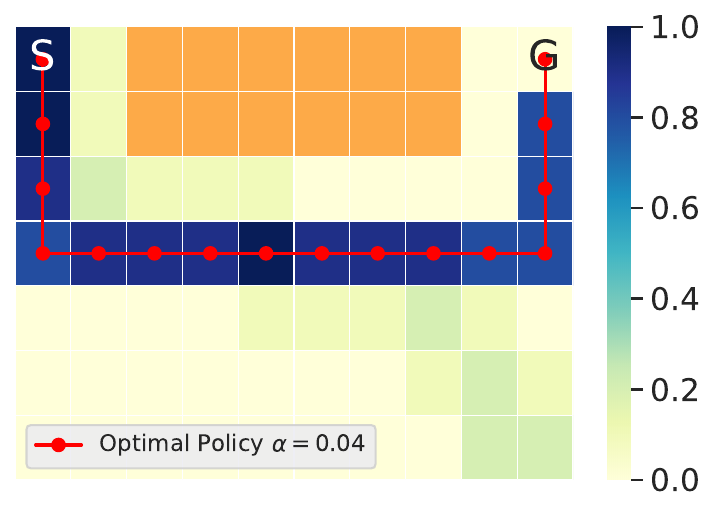}
    \caption{ACReL$_{0.04}$ ($\eta=25$)}
    \label{fig:averaged_seeds_1005}
    \includegraphics[width=\textwidth]{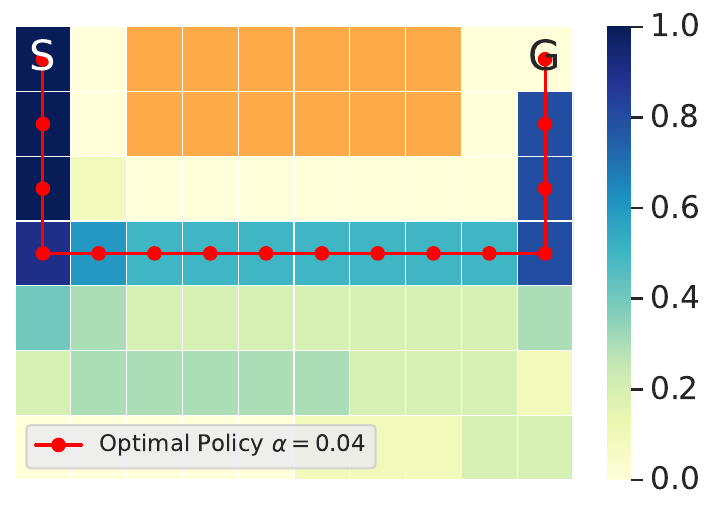}
    \caption{IQN-CVaR$_{0.04}$}
\end{subfigure}\hfill
\begin{subfigure}[t]{0.3 \linewidth}
    \includegraphics[width=\textwidth]{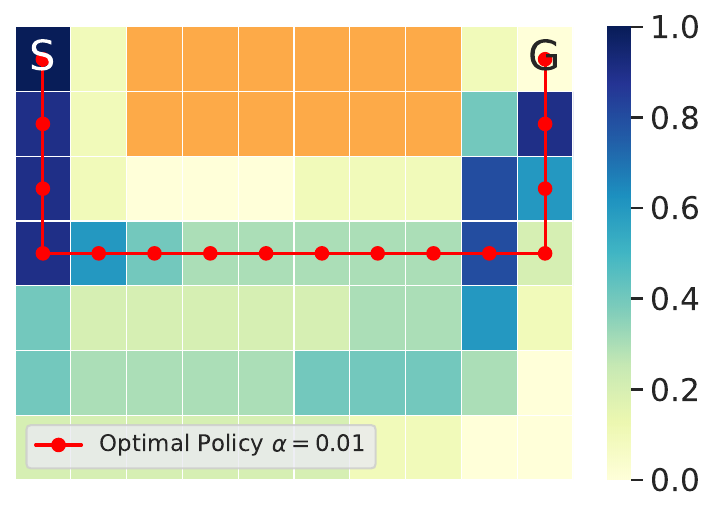}
    \caption{ACReL$_{0.01}$ ($\eta=100$)}
    \label{fig:averaged_seeds_2005}
    \includegraphics[width=\textwidth]{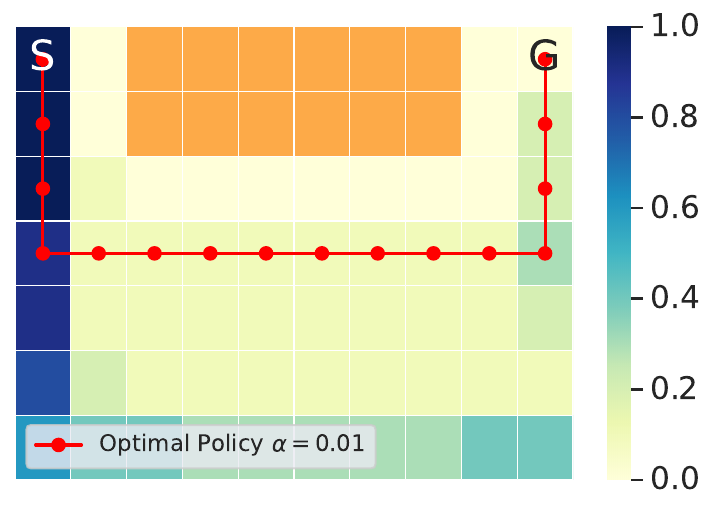}
    \caption{IQN-CVaR$_{0.01}$}
\end{subfigure}
\caption{Agent policies learned by ACReL and IQN-CVaR for different CVaR$_\alpha$ objectives, $\alpha \in \{1, 0.04, 0.01\}$ in an envrionment with probability of random action $p=0.05$.}
\label{fig:averaged_seeds005}
\end{figure*}

Red trajectories in Figure\ref{fig:averaged_seeds005} correspond to the resolution of the environment with CVaR Policy Iteration. 
As expected, Figure~\ref{fig:averaged_seeds_0005} shows that in the absence of risk-sensitivity ($\alpha=1$), the optimal behavior is to follow the cliff by taking the shortest path to the goal. However, in the presence of risk-sensitivity ($\alpha<1$), Figures~\ref{fig:averaged_seeds_1005} and~\ref{fig:averaged_seeds_2005} show that the the optimal behavior is a careful path that deviates from the lava. However, we notice that the optimal behavior corresponding to a CVaR$\_{0.01}$ and CVaR$\_{0.04}$ are the same. This is because the stochasticity of the environment is low ($p=0.05$), therefore the risk of accidentally falling into the lava is lower.

Blue trajectories in Figure~\ref{fig:averaged_seeds005} display the learned agent policy averaged over the 10 seeds for each of the 3 considered $\text{CVaR}_{\alpha}$ optimization tasks. The Figure confirms that our adversarial learning method results in a safer policy as well. 
Furthermore, the averaged behavior is consistent over 10 different randomness realizations. The noticeable noise in the averaged agent policy displayed in Figure~\ref{fig:averaged_seeds005} is due to punctual convergence instabilities. Despite the instabilities inherent to the training procedure, the agent trained with ACReL consistently displayed convergence for the 3 considered tasks. IQN-CVaR converges successfully for $\text{CVaR}_{1}$ and $\text{CVaR}_{0.04}$ but fails to resolve the environment for $\text{CVaR}_{0.01}$

\end{document}